\documentclass[nohyperref]{article}

\usepackage[dvipsnames]{xcolor}

\usepackage{microtype}
\usepackage{graphicx}
\usepackage{subfigure}
\usepackage{booktabs} 
\usepackage{amsmath}
\usepackage{appendix}
\usepackage{float}

\usepackage[pagebackref=true, colorlinks=true,urlcolor=blue,citecolor=black,linkcolor=blue,anchorcolor=black]{hyperref}

\usepackage[arxiv]{icml2022}

\usepackage{amsmath}
\usepackage{amssymb}
\usepackage{mathtools}
\usepackage{amsthm}

\usepackage[capitalize,noabbrev]{cleveref}

\DeclareMathOperator*{\argmax}{argmax}

\definecolor{zgray}{RGB}{0,187,214} 
\definecolor{zgray}{RGB}{225,0,124} 
\definecolor{zgray}{RGB}{239,164,0} 
\definecolor{zgray}{RGB}{34,139,34} 
\definecolor{zgray}{RGB}{70,70,70} 

\definecolor{zgreen}{RGB}{34,139,34} 

\newcommand{\airfoil}[0]{\textsc{Airfoil}}
\newcommand{\twodfluids}[0]{\textsc{2D Fluid Tools}}
\newcommand{\threedfluids}[0]{\textsc{3D Watercourse}}
\newcommand{\mgn}[0]{\textsc{MeshGraphNets}}
\newcommand{\dafoam}[0]{DAFoam}
\newcommand{\contain}[0]{\textit{Contain}}
\newcommand{\ramp}[0]{\textit{Ramp}}
\newcommand{\maze}[0]{\textit{Maze}}
\newcommand{\direction}[0]{\textit{Direction}}
\newcommand{\twopools}[0]{\textit{2 Pools}}
\newcommand{\threepools}[0]{\textit{3 Pools}}

\renewcommand*{\backref}[1]{}
\renewcommand*{\backrefalt}[4]{%
    \ifcase #1%
          \or Cited on page~#2.%
    \else Cited on pages~#2.%
\fi%
}

\usepackage[disable,textsize=tiny]{todonotes}

\icmltitlerunning{Physical Design using Differentiable Learned Simulators}

\begin{document}

\twocolumn[
\icmltitle{Physical Design using Differentiable Learned Simulators}

\icmlsetsymbol{equal}{*}

\begin{icmlauthorlist}
\icmlauthor{Kelsey R. Allen}{equal,DeepMind}
\icmlauthor{Tatiana Lopez-Guevara}{equal,DeepMind}
\icmlauthor{Kimberly Stachenfeld}{equal,DeepMind}
\icmlauthor{Alvaro Sanchez-Gonzalez}{DeepMind}
\icmlauthor{Peter Battaglia}{DeepMind}
\icmlauthor{Jessica Hamrick}{DeepMind}
\icmlauthor{Tobias Pfaff}{DeepMind}
\end{icmlauthorlist}

\icmlaffiliation{DeepMind}{DeepMind, UK}

\icmlcorrespondingauthor{Kelsey R. Allen}{krallen@deepmind.com}
\icmlcorrespondingauthor{Tatiana Lopez-Guevera}{zepolitat@deepmind.com}
\icmlcorrespondingauthor{Kimberly Stachenfeld}{stachenfeld@deepmind.com}
\icmlcorrespondingauthor{Tobias Pfaff}{tpfaff@deepmind.com}

\icmlkeywords{Machine Learning, ICML}

\vskip 0.3in
]

\printAffiliationsAndNotice{Equal contribution. Authors listed alphabetically.} 

\begin{abstract}
Designing physical artifacts that serve a purpose---such as tools and other functional structures---is central to engineering as well as everyday human behavior.
Though automating design has tremendous promise, general-purpose methods do not yet exist.
Here we explore a simple, fast, and robust approach to inverse design which combines learned forward simulators based on graph neural networks with gradient-based design optimization.
Our approach solves high-dimensional problems with complex physical dynamics, including designing surfaces and tools to manipulate fluid flows and optimizing the shape of an airfoil to minimize drag.
This framework produces high-quality designs by propagating gradients through trajectories of hundreds of steps, even when using models that were pre-trained for single-step predictions on data substantially different from the design tasks.
In our fluid manipulation tasks, the resulting designs outperformed those found by sampling-based optimization techniques.
In airfoil design, they matched the quality of those obtained with a specialized solver.
Our results suggest that despite some remaining challenges, machine learning-based simulators are maturing to the point where they can support general-purpose design optimization across a variety of domains.
\end{abstract}

\section{Introduction}
 
Humans are creators.
Our ancestors created stone tools which led to innovations in hunting and food consumption, aqueducts and irrigation systems which revolutionized farming and urban habitation, and more recently, airplanes which let us cross the globe in hours.
Automatically designing objects to exhibit a desired property---often referred to as \emph{inverse design}---promises to transform science and engineering, including aerodynamics \cite{eppler2012airfoil}, material design \cite{butler2016computational}, optics \citep{colburn2021inverse}, and robotics \cite{Gupta2021, xu2021diffsim}.

Despite its promise, widespread practice of inverse design has been limited by the availability of fast, general-purpose simulators.
In science and engineering, many methods rely on specialized ``classical'' solvers, which are handcrafted to simulate a particular physical process.
While accurate and reliable, these solvers can be quite slow, may not provide gradients, and are narrow in their applicability \citep{cranmer2020frontier}.
In robotics and reinforcement learning, simulators are often learned, but accumulate errors over long time horizons and often struggle to generalize beyond their training data \citep{janner2019mbpo,talvitie2014model,venkatraman2015improving}, making them unsuitable for design optimization without further finetuning.

Recently, a class of learned physics simulators based on graph neural networks (GNNs) has been proposed \citep{pfaff2021learning,sanchez2020learning}.
These models have shown success in general-purpose physical prediction, exhibiting high accuracy and generalization ability.
However, this may still be insufficient for inverse design problems, as optimizers can exploit regions in the state space where model predictions are unreliable \citep{lutter2021learning}. Models must therefore be more than just accurate overall: they must also be robust and smooth.
It is unknown whether GNN-based physics simulators exhibit these properties.

\begin{figure*}[!t]
\centering
\includegraphics[width=0.96\textwidth]{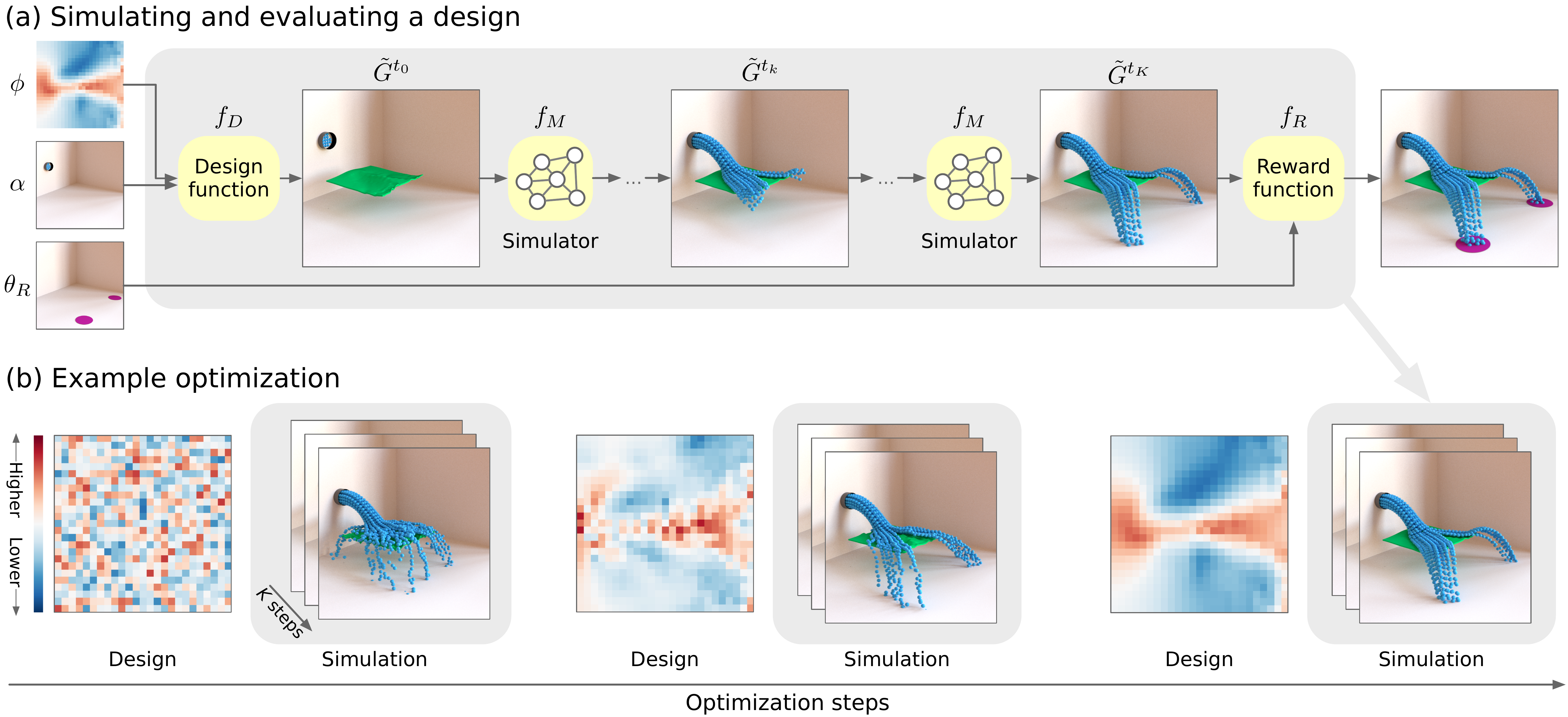}
\vspace{-0.25em}
\caption{Optimizing a physical design. Here, the goal is to direct a stream of water (shown in blue) into two ``pools'' (shown in purple) by designing a ``landscape'' (shown in green) parameterized as a 2D height field. (a) The simulation pipeline takes in a design $\phi$ and initial conditions $\alpha$ and uses the design function $f_D$ to produce an initial state. The simulation is rolled out with a pre-trained learned simulator $f_M$ for $K$ steps, at which point the final state is passed (along with reward parameters $\theta_R$) to the reward function $f_R$, which computes the quality of the design. (b) Each step of optimization involves rolling out the simulation and then adjusting the design ($\phi$) accordingly using an optimizer such as gradient descent or CEM. Shown are selected frames from gradient-based optimization in the \twopools{} task of the \threedfluids{} domain.}
\vspace{-0.5em}
\label{fig:schematic}
\end{figure*}

In this paper, we optimize physical designs by performing gradient descent through pretrained, GNN-based, state-of-the-art learned simulators.
We use this approach to perform successful inverse design without requiring further finetuning of the simulator.
Across two high-dimensional fluid manipulation tasks (\twodfluids{} and \threedfluids{}) and a design task from aerodynamics (\airfoil{}), we show that learned simulators:
(1) produce high-quality designs across diverse physical tasks with complex particle- or mesh-based physics, while using the same underlying GNN architecture;
(2) generalize sufficiently to permit designs far outside their training data;
(3) support gradient-based optimization over hundreds of time steps, through states with thousands of particles, in tasks with up to 625 design parameters (and as a result, produce better designs than sampling-based optimization using a classical simulator); and
(4) can be much faster than specialized simulators used in engineering, while generating designs of similar quality.
Overall, our results are a proof-of-concept for how state-of-the-art learned simulators can be used at scale to optimize designs for different physical tasks.
\section{Background}  
\label{sec:background} 
Solving inverse problems with physical simulators has a long history in science and engineering, spanning data assimilation for weather modeling \citep{navon2009data}, system identification in robotics \citep{guevara2017adaptable, seita_fabrics_2020}, and tomographic and geophysical imaging \citep{cui2016review, pascual1999review}.
Inverse design can be framed as an inverse problem in which the objective is to optimize design parameters to produce some desired target property.
Simulation-based inverse design has been studied in a variety of disciplines, including nanophotonics \citep{molesky2018inverse}, material science \citep{Dijkstra2021predictive}, mechanical design \cite{coros2013computational}, and aerodynamics \cite{anderson1999aerodynamic, Rhie_1983}.

Classical numerical solvers used for inverse design can be highly accurate, but are often inefficient (making sampling-based inference methods infeasible for high-dimensional designs) and domain-specialized ( prohibiting general-purpose inverse design across domains \citep{choi2021use}). 
Differentiable simulators \citep{freeman2021brax,hu2019difftaichi,Schenck2018SPNetsDF} have recently garnered attention, as they allow for more sample-efficient gradient-based optimization. 
However, like classical solvers, they are still typically narrow in application scope, as many simulation techniques are hard to express as a differentiable program (e.g. constraint dynamics and multiphysics coupling).

The last few years have seen increased interest in using machine learning to accelerate inverse design across a variety of applications \citep[e.g.][]{Challapalli2021inverse,Christensen2020predictive,Bombarelli2018Automatic,Forte2022Inverse,
Hoyer2019neural,Kumar2020Inverse,Li2020efficient,Liu2018Generative,Sha2021Machine,ZHENG2021113894}.
These methods can provide impressive speedups over classical approaches by using learned generative models to propose designs (thus calling an expensive simulator fewer times), or by learning a scoring function which maps designs to target values (replacing the simulator entirely). However, they are based on components with limited out-of-domain generalization,
restricting new designs to configurations near the training data or requiring model refinement.

We instead propose to replace classical simulators with learned simulators.
Like learned scoring functions, learned simulators can be faster than classical simulators \citep{kochkov2021machine,stachenfeld2021learned} and are differentiable when parameterized as a neural network.
Furthermore, learned simulators mimic the underlying physical dynamics independent of the design task and are therefore more likely to generalize.
Physics simulators have been successfully implemented as learned, differentiable models of complex dynamics such as fluids, rigid-body interactions, and soft-body systems \citep{bhatnagar2019prediction,li2020fourier,mrowca2018,Rudy2017data,thuerey2020deep,ummenhofer2020lagrangian,wang2020physicsinformed}. 
Graph-based models in particular are promising candidates for design problems, having demonstrated high accuracy, stability, efficiency, and generalization performance \citep{belbute2020combining, pfaff2021learning,sanchez2020learning}.

However, high-quality forward models do not necessarily translate into better downstream task performance \cite{hamrick2020role,lutter2021learning}. While learned simulators have been used successfully for planning and control in low-dimensional state spaces \citep[up to 21 degrees of freedom (DOF), e.g.][]{bharadhwaj2020model,sanchez2018graph,wang2019benchmarking} or more complex domains with small action spaces \citep[6 DOF, e.g.][]{li2018learning}, 
they require replanning at every timestep to avoid error accumulation. It is not known whether learned simulators can support gradient-based, high-dimensional inverse design which demands high accuracy, well-behaved gradients, long-term rollout stability, and generalization beyond the training data.

Here we study inverse design in non-rigid, graph-based physical systems with up to 625 design dimensions and 2000 state dimensions, over 50--300 timesteps, without replanning after the design period. We show that using gradient-based optimization with learned, general-purpose simulators is an effective choice for inverse design.
\section{Problem Formulation}

\begin{figure*}[h!]
\centering
\includegraphics[width=0.48\textwidth]{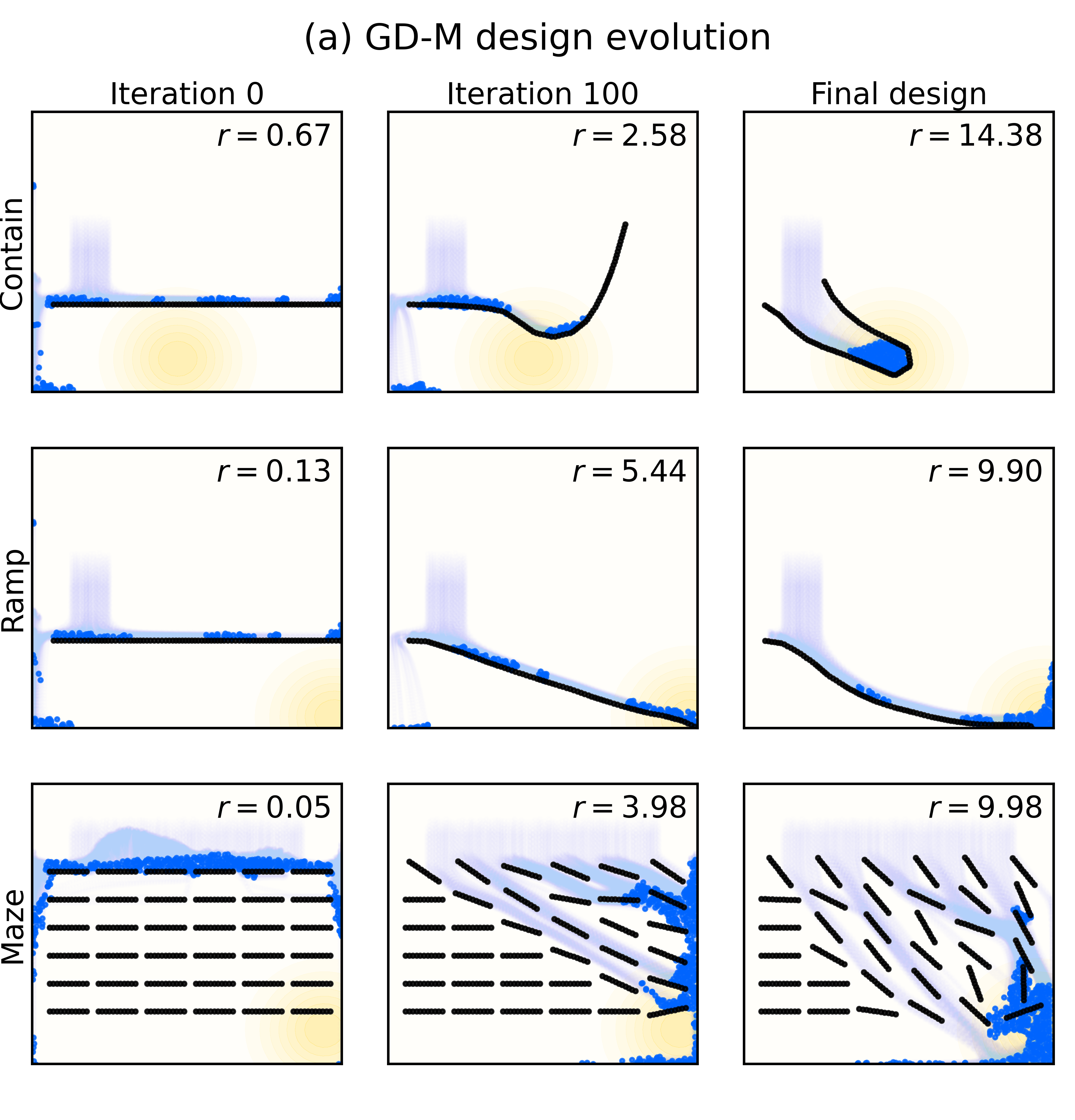}%
\hspace{1em}\includegraphics[width=0.16\textwidth]{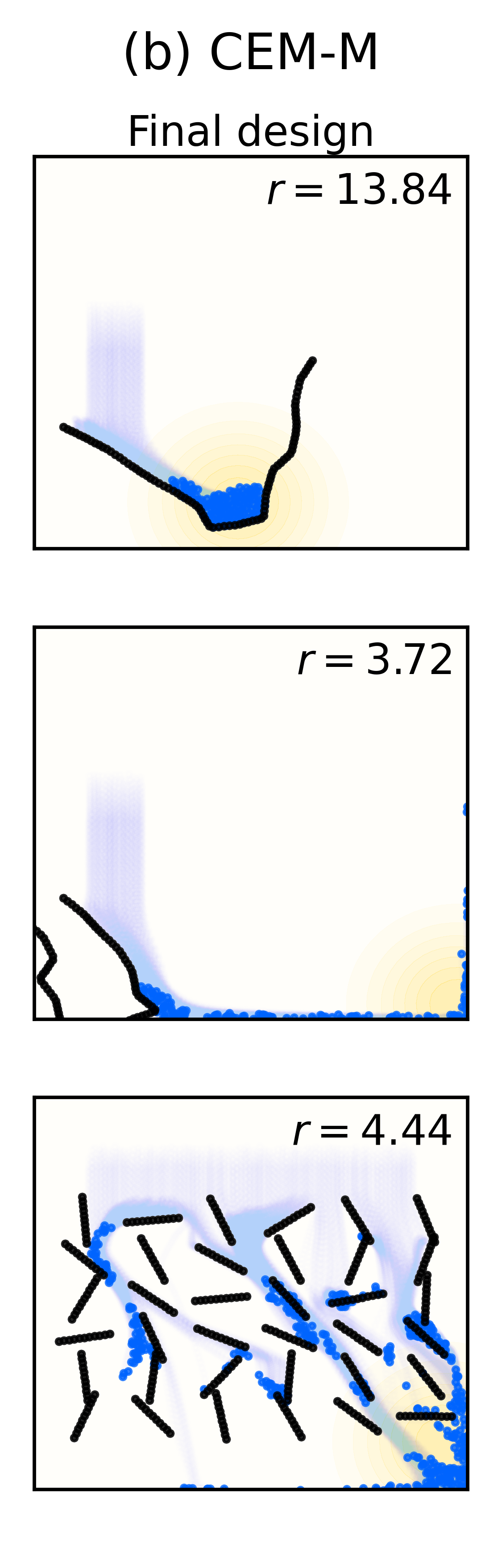}%
\hspace{1em}\includegraphics[width=0.16\textwidth]{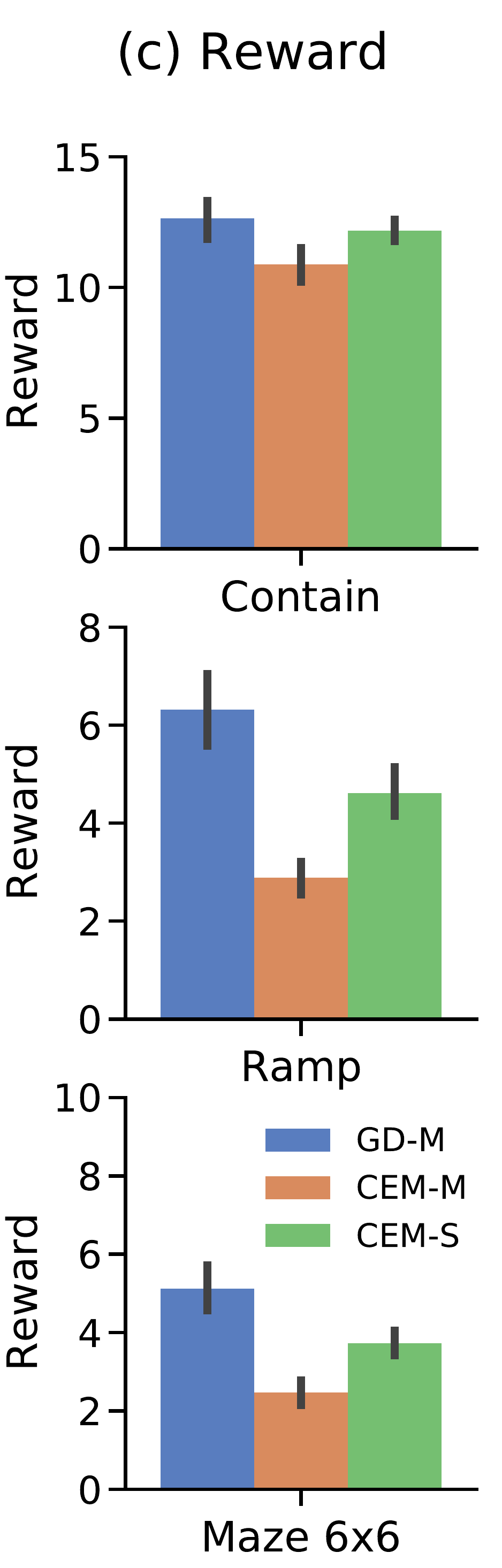}
\vspace{-0.25em}
\caption{\twodfluids{} results. The state spaces consist of $10^2$--$10^3$ particles and the design spaces of 16--36 parameters. (a) Evolution of designs found by GD-M during optimization for each \twodfluids{} task. Visualizations correspond to simulations of the designs under $f_S$. The design is shown in black, fluid particles in blue, and Gaussian reward in yellow. The transparent particles show the location of fluid for $t<t_K$, and the solid particles show the location of fluid at the final frame $(K=150)$. $r$ denotes reward for the current design.
(b) Final designs found by CEM-M.
(c) Mean reward over 50 reward locations (with bootstrapped $95\%$ confidence intervals) obtained by each optimizer across the \twodfluids{} tasks.  For \contain{} and \ramp{}, results are shown for 16 joints; for \maze{}, results are shown for a $6\times6$ grid of 36 rotors. Across these tasks, GD-M outperforms both CEM-M and CEM-S.}
\vspace{-0.5em}
\label{fig:2dfluids_results}
\end{figure*}

Consider the design task depicted in \autoref{fig:schematic}, in which the goal is to direct a stream of water (shown in blue) into two ``pools'' (shown in purple) by designing a ``landscape'' (shown in green) parameterized as a 2D height field.
Here, an ideal design will create ridges and valleys that direct fluid into the two targets.
In the next sections, we formalize what it means to find and evaluate such a design and discuss our choices for simulator and optimizer.

\subsection{Learned simulators}

To demonstrate the utility of learned simulators for finding physical designs, we rely on the recently developed \mgn{} model \citep{pfaff2021learning}, which is an extension of the GNS model for particle simulation \citep{sanchez2020learning}.
\mgn{} is a type of message-passing graph neural network (GNN) that performs both edge and node updates \citep{battaglia2018relational,gilmer2017neural}, and which was designed specifically for physics simulation.
Here, we briefly summarize how the learned simulator works, and refer interested readers to the original papers for details.

We consider simulations over physical states represented as graphs $G\in\mathcal{G}$.
The state $G = (V, E)$ has nodes $V$ connected by edges $E$, where each node $v\in V$ is associated with a position $\mathbf{u}_v$ and additional dynamical quantities $\mathbf{q}_v$. 
These graphs may be either meshes (as in \mgn{}) or particle systems (as in GNS).
In a mesh-based system (such as \airfoil{}), $V$ and $E$ correspond to vertices and edges in the mesh, respectively.
In a particle system (such as \twodfluids{}), each node corresponds to a particle and edges are computed dynamically based on proximity.
Under this framework, we can also consider hybrid mesh-particle systems (such as \threedfluids{}).
See \autoref{app:model} for model implementation details, and \autoref{app:domains} for details on the representation used for each domain.

The simulation dynamics are given by a ``ground-truth'' simulator $f_S:\mathcal{G}\rightarrow\mathcal{G}$ which maps the state at time $t$ to that at time $t+\Delta t$.
The simulator $f_S$ can be applied iteratively over $K$ time steps to yield a trajectory of states, or a ``rollout,'' which we denote $(G^{t_0}, ..., G^{t_K})$.
Using \mgn{}, we learn an approximation $f_M$ of the ground-truth simulator $f_S$.
The learned simulator $f_M$ can be similarly applied to produce rollouts $(\tilde{G}^{t_0}, \tilde{G}^{t_1},  ..., \tilde{G}^{t_K})$, where $\tilde{G}^{t_0} = G^{t_0}$ represents initial conditions given as input.
We note that a learned simulator allows us to take much larger time-steps than $f_S$, permitting shorter rollout lengths: in our running example, one model step corresponds to 200 internal steps of the classical simulator.
See \autoref{fig:schematic}a for an illustration of simulation using a learned model.

\subsection{Optimizing design parameters}
\label{sec:optimization}

To optimize a physical design, we leverage the pipeline shown in \autoref{fig:schematic}: (1) transform design parameters into an initial scene, (2) simulate the scene using $f_M$ or $f_S$, (3) evaluate how well the simulation achieves the desired behavior, and (4) adjust the design parameters accordingly.

\vspace{-0.5em}
\paragraph{Design parameters}
To produce the initial state $G^{t_0}$, we introduce a differentiable design function $f_D: \Phi\times \mathcal{A} \rightarrow \mathcal{G}$ which maps design parameters $\phi\in\Phi$ and other initial conditions $\alpha\in \mathcal{A}$ to an initial state: $G^{t_0}=f_D(\phi, \alpha)$.
In our landscape design task (\autoref{fig:schematic}), $\phi$ is the 2D height field of the mesh, while $\alpha$ is the non-controllable objects in the scene like the initial position of the fluid.

\vspace{-0.5em}
\paragraph{Maximizing reward}
The reward function $f_R: \mathcal{G}\times\Theta_R\rightarrow\mathbb{R}$ maps the final state of a length-$K$ trajectory ($G^{t_K}$ or $\tilde{G}^{t_K}$) and parameters $\theta_R\in\Theta_R$ to a scalar value.
In our running example, the reward function is defined as the Gaussian likelihood of each fluid particle under the closest ``pool'', averaged across particles (\autoref{fig:schematic}a).

We define the full objective under the ground-truth simulator $f_S$ as $J_S(\phi):=f_R(f_S^{(K)}(f_D(\phi, \alpha)); \theta_R)$, where $f_S^{(K)}$ indicates $K$ applications of the simulator.
We want to find the design parameters that maximize $J_S$, i.e. $\phi^* = \argmax_\phi J_S(\phi)$.
We can approximate this optimization using a learned simulation model instead by maximizing $J_M(\phi):=f_R(f_M^{(K)}(f_D(\phi, \alpha)); \theta_R)$.

\looseness=-1
\paragraph{Optimizers}
Optimal design parameters $\phi^*$ can be found using any generic optimization technique.
Given the differentiability of the learned simulator $f_M$, we are particularly interested in evaluating gradient-based optimization, which requires fewer function evaluations and scales better to large design spaces than sampling-based techniques \citep{bharadhwaj2020model}.
We focus on the Adam optimizer \citep{kingma:adam}, which we use to find $\phi^*$ by computing the gradient $\nabla_\phi J_M(\phi)$.
This involves backpropagating gradients through the reward function $f_R$, length-$K$ rollout produced by $f_M^{(K)}$, and design function $f_D$.

As a baseline, we consider the cross-entropy method (CEM) \citep{rubinstein2004cem}, a gradient-free sampling-based technique that is popular in model-based control \cite{chua2018deep,wang2019benchmarking}.
CEM can be used with any simulator and works by sampling a population of candidates for $\phi$ and evolving them to maximize the reward.
However, CEM requires multiple evaluations of $f_M$ or $f_S$ per optimizer step (depending on the population size, which is 20--40), whereas Adam considers only a single candidate $\phi$ and thus only a single evaluation per step.

Across our design tasks (\autoref{sec:design-tasks}) we compared: gradient descent with the learned simulator (\textbf{GD-M}), CEM with the learned simulator (\textbf{CEM-M}), and CEM with the ground-truth simulator (\textbf{CEM-S}).
In all tasks, $f_S$ is non-differentiable, preventing a comparison to GD-S.
However, in the special case of \airfoil{}, we compare to \textbf{\dafoam{}} \cite{he2020dafoam}, a specialized solver which computes gradients with the adjoint method.

\begin{figure*}[!h]
\centering
\includegraphics[width=0.96\textwidth]{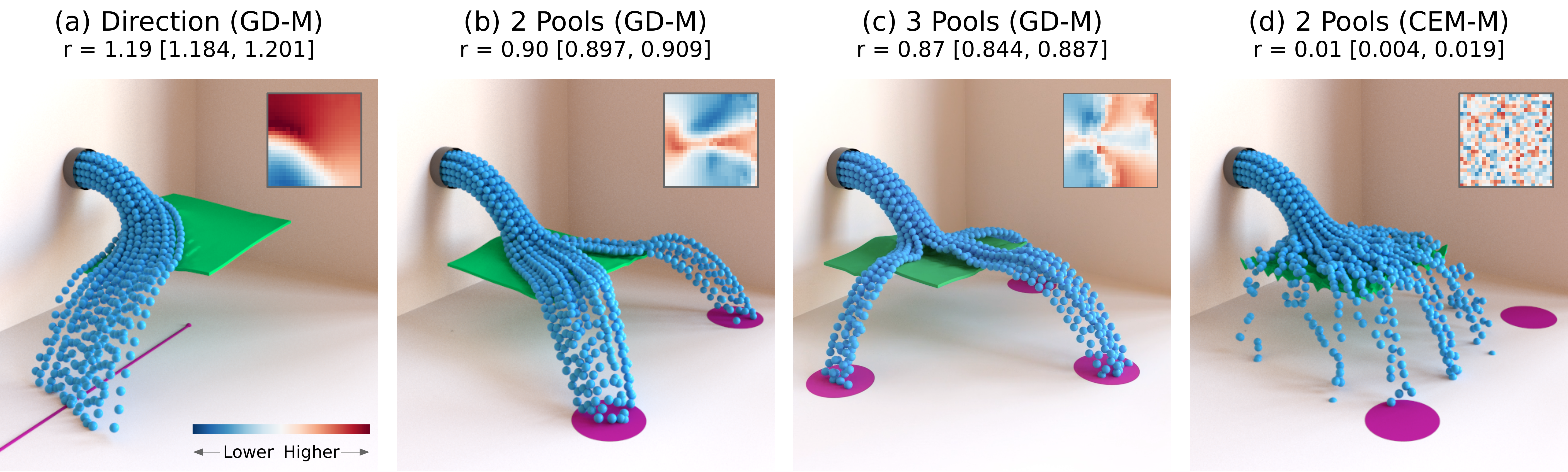}
\vspace{-0.25em}
\caption{\threedfluids{} results found with GD-M or CEM-M and evaluated with $f_S$. Simulations use up to 2000 particles and 625 design parameters. The heightmap of a 2D landscape is optimized to redirect the fluid towards the purple targets; birds eye views of heightmaps  are shown in the upper right corner of each subplot. In this high-dimensional task domain, GD finds designs with high reward (a-c), while CEM fails to find meaningful solutions for \twopools{} (d) as well as the other tasks (\autoref{fig:accuracy_comparison}a).
Each subplot reports the mean reward (r) and bootstrapped [lower, upper] 95\% confidence intervals for the corresponding optimizer and task (averaged over 10 randomized initial designs for each task variation, see \autoref{app:3dfluids}).
}
\vspace{-0.5em}
\label{fig:3dfluids}
\end{figure*}

\looseness=-1
\paragraph{Evaluation}
Unless otherwise noted, we always evaluate the quality of an optimized design $\phi^*$ using the ground-truth objective $J_S(\phi^*)$, regardless of whether $\phi^*$ was found using the learned model $f_M$ (as in CEM-M and GD-M) or the ground-truth simulator $f_S$ (as in CEM-S).
We also use rollouts from $f_S$ to produce visualizations in the figures.

\section{Design Tasks}
\label{sec:design-tasks}
We formulated a set of design tasks across three different physical domains with high-dimensional state spaces and complex dynamics. Each domain uses a different ground-truth simulator $f_S$, which is used to evaluate designs and pre-train the learned simulator model $f_M$.

\subsection{\twodfluids{}}
\label{sec:2d-fluid-design}

Inspired by existing 2D physical reasoning benchmarks \citep{allen2019tools,bakhtin2019phyre}, these tasks involve creating one or more 2D ``tool'' shapes to direct fluid into a particular goal region (\autoref{fig:2dfluids_results}, \autoref{app:2dfluids}).
We consider three tasks with 4--48 design parameters where the goal is to guide the fluid such that each particle comes as close as possible to the center of a randomly-sampled yellow reward region.
In \contain{}, the joint angles $\phi_\mathrm{joints}$ of a multi-segment tool must be optimized to catch the fluid by creating cup- or spoon-like shapes.
In \ramp{}, the joint angles $\phi_\mathrm{joints}$ of a multi-segment tool must be optimized to guide the fluid to a distant location.
In \maze{}, the rotation $\phi_\mathrm{rot}$ of multiple tools must be optimized to funnel the fluid to the target location.
Fluid dynamics are represented using $10^2$--$10^3$ particles, and unrolled for up to 300 time steps.

The learned model $f_M$ is trained on the 2D WaterRamps dataset released by \citet{sanchez2020learning} which uses the solver in \citet{hu2018moving} to generate trajectories.
The dataset contains scenes with 1--4 straight line segments; no curved lines or large numbers of obstacles are shown.
Therefore, designs which solve \twodfluids{} tasks are necessarily far out of distribution for the learned model.

\subsection{\threedfluids{}}
\label{sec:3d-fluid-design}

To evaluate higher-dimensional inverse design, we created a ``landscaping'' task that requires optimizing a 3D surface to guide fluids into different areas of an environment (\autoref{fig:3dfluids}, \autoref{app:3dfluids}).
Specifically, water flows out of a pipe and onto an obstacle parameterized by $\phi_\mathrm{map}$, a $25 \times 25$ heightmap (625 design parameters).
In \direction{}, $\phi_\mathrm{map}$ is optimized to redirect the fluid stream towards a specified direction.
In \twopools{} and \threepools{}, $\phi_\mathrm{map}$ is optimized to split the fluid stream such that particles hitting the floor land as close as possible to one of two or three specified pools.
Each task has multiple variants, such as different target directions in \direction{}.
The simulation is unrolled for 50 time steps, and contains up to 2048 particles.
The learned model $f_M$ is trained on data generated from the simulator in \citet{bender2015divergence}.

\begin{figure*}[h!]
\centering
\includegraphics[width=0.24\textwidth]{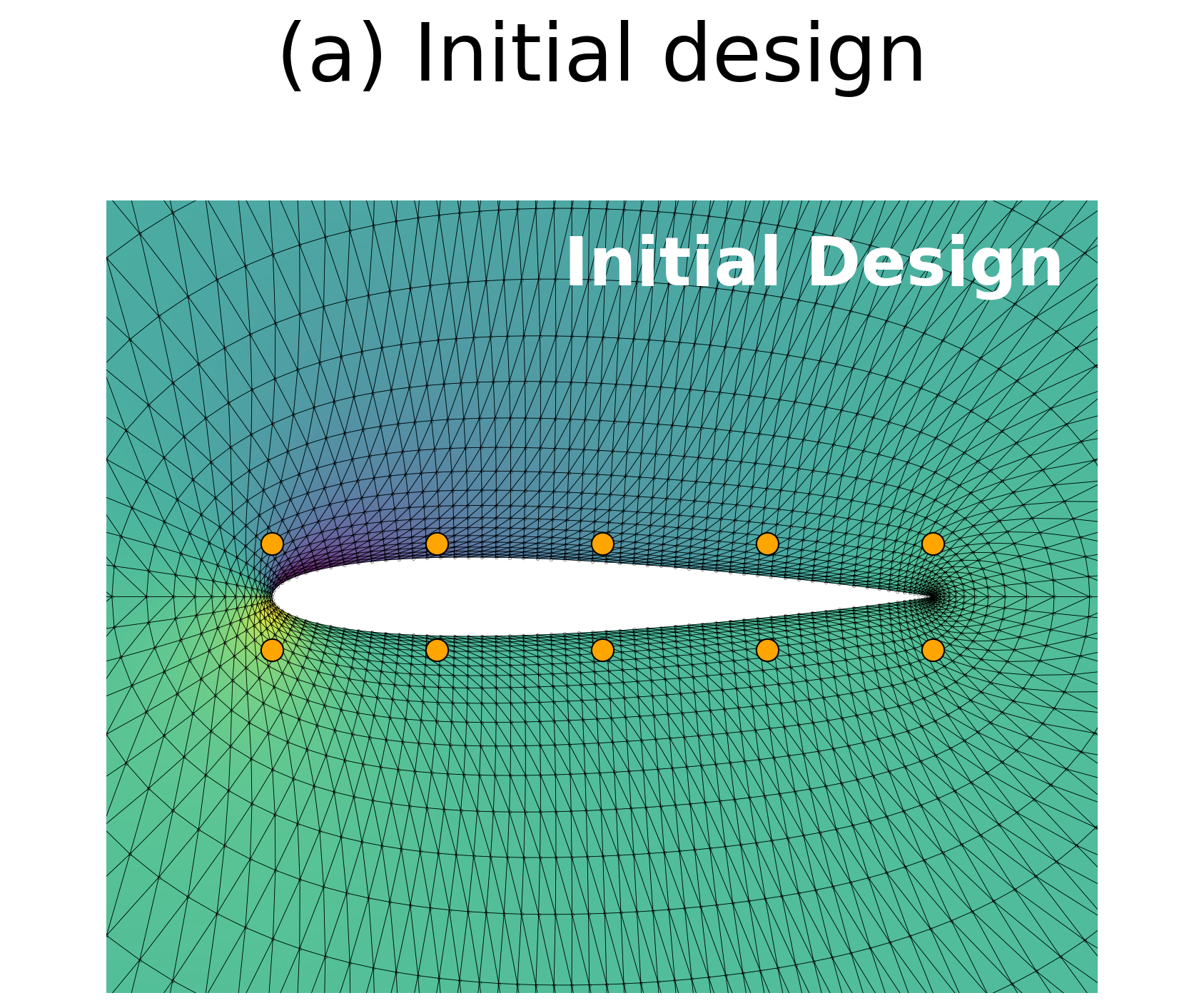}%
\includegraphics[width=0.24\textwidth]{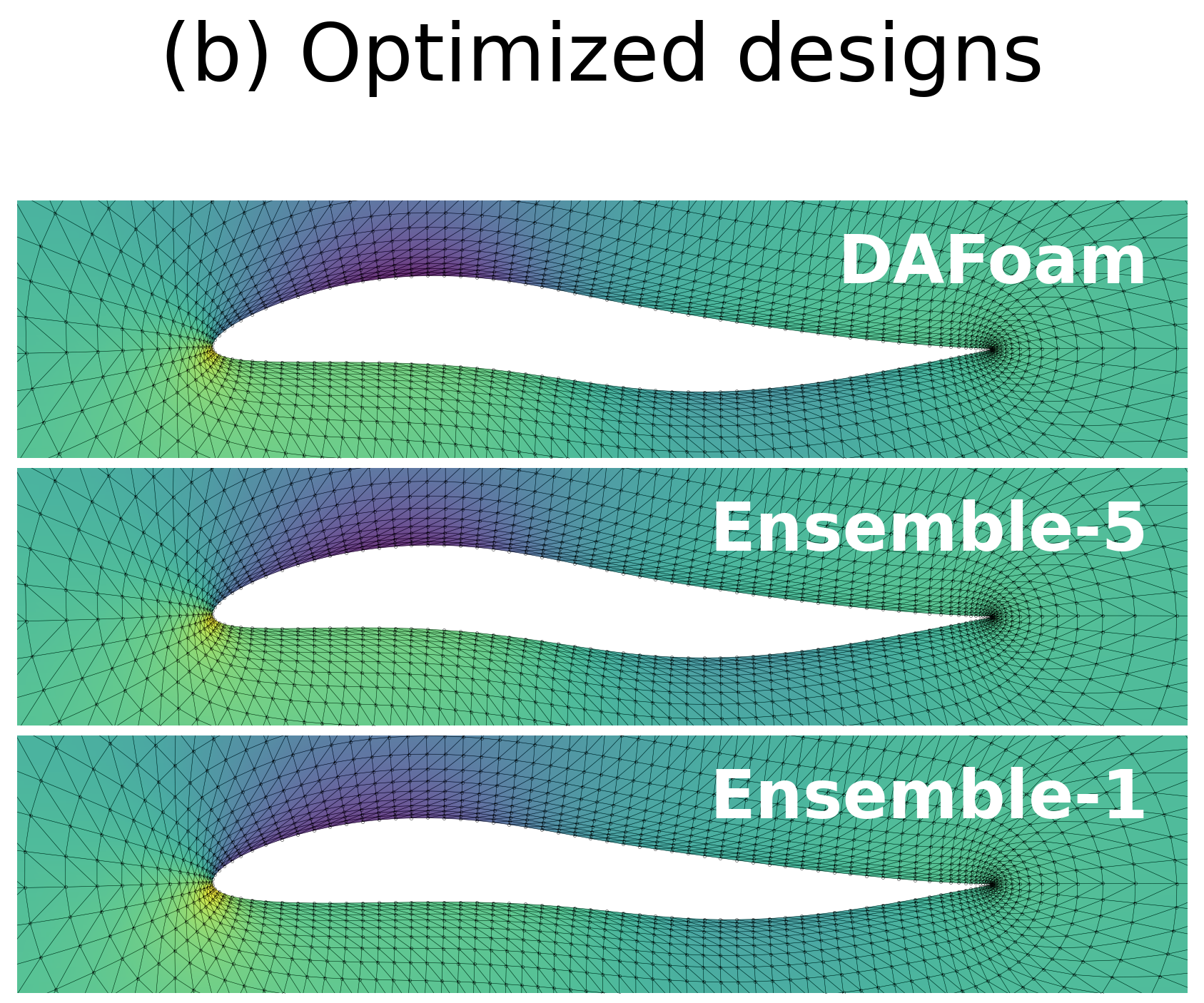}%
\includegraphics[width=0.24\textwidth]{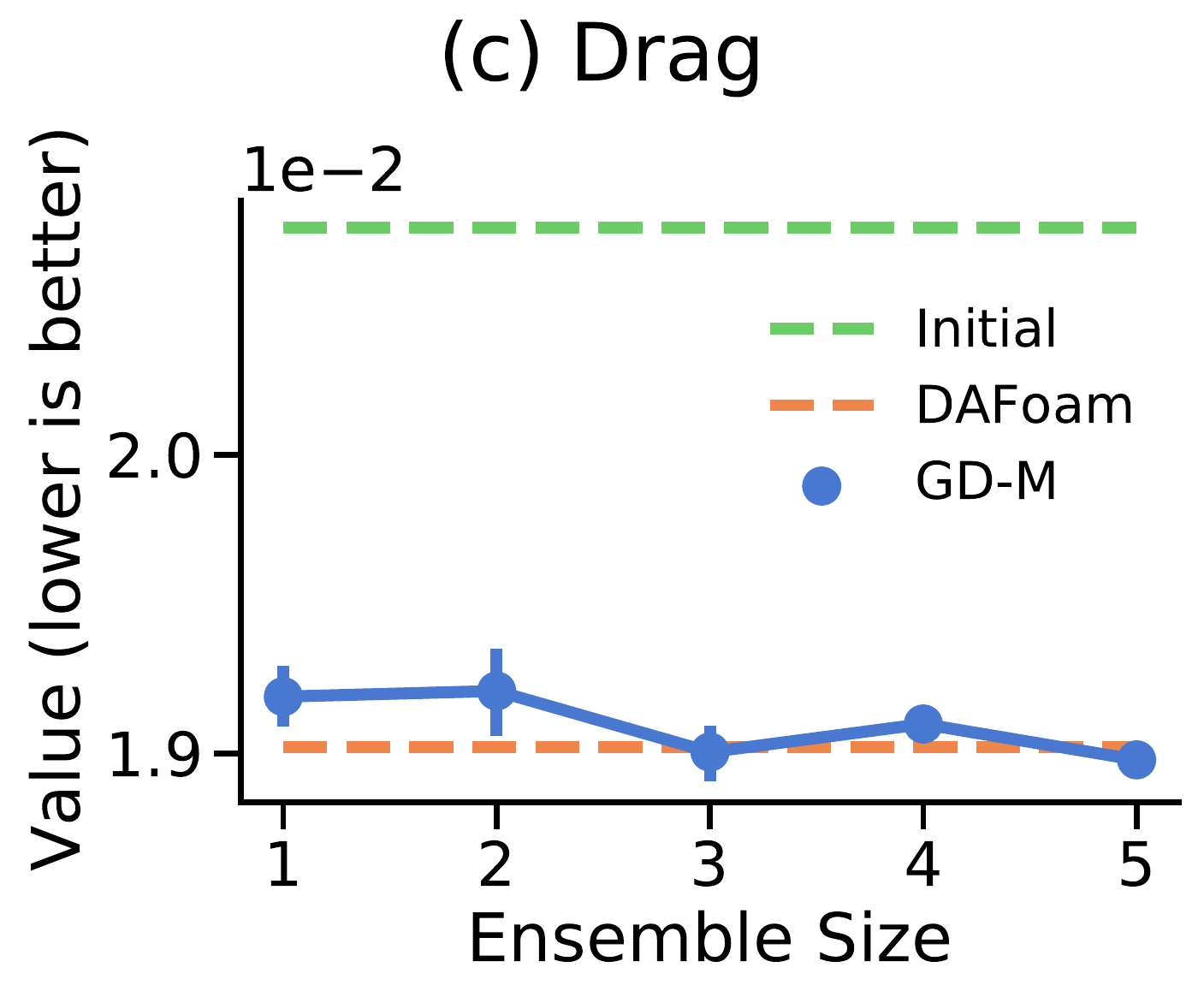}%
\includegraphics[width=0.24\textwidth]{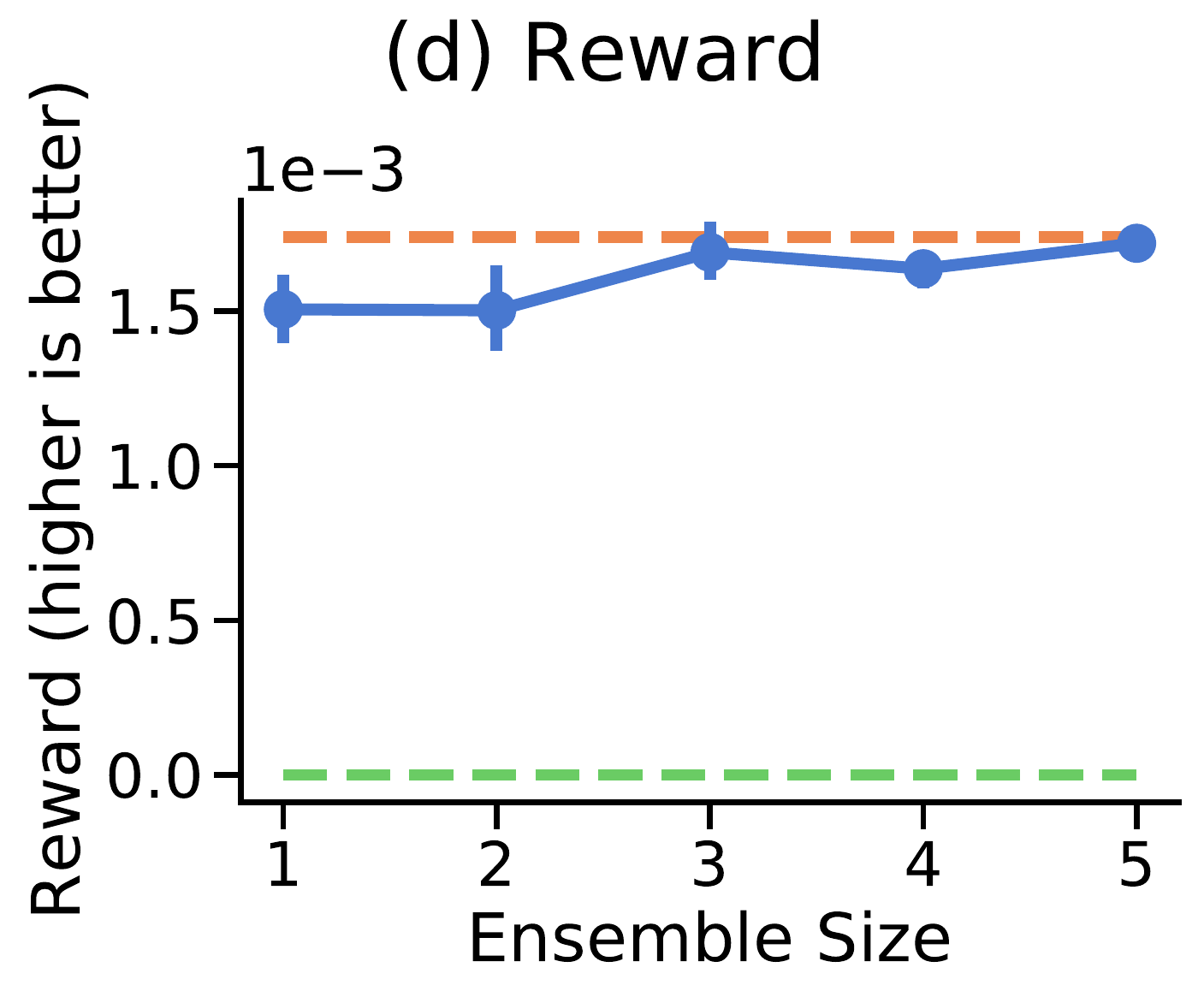}
\vspace{-0.25em}
\caption{\airfoil{} results. (a) An initial airfoil design is warped by moving 10 control points (orange dots). The physics model simulates the resulting aerodynamics on a 4158 node mesh, based on which lift and drag are computed. (b) For the task of finding a minimum-drag configuration under constant lift constraint, gradient-based learned design is able to find similar designs to specialized solver \dafoam{}, both using single models and ensembles. (c-d) Larger ensemble sizes of 3--5 achieve a close quantitative match to \dafoam{} for both drag and overall reward.
Shown are means over 10 randomized initial designs, with bootstrapped 95\% confidence intervals.}
\vspace{-0.5em}
\label{fig:airfoil}
\end{figure*}

\subsection{\airfoil{}}
\label{sec:airfoil}

Shape optimization in aerodynamics is one area where gradient-based optimization is routinely applied using traditional simulators \citep{buckley2010airfoil}.
Here, we consider the well-studied task of drag optimization of a 2D airfoil profile (\autoref{fig:airfoil}, \autoref{app:airfoil}).
In this task, a wing is defined using a curve on a 2D mesh, which can be deformed using a set of 10 control points $\phi_\textrm{ctrl}$.
The reward function is formulated as optimizing the wing shape to minimize drag under certain constraints such as constant lift and bounds on the shape to prevent degenerate (e.g. infinitely thin) configurations. 
Lift and drag coefficients are computed by running an aerodynamics simulation on a 4158 node mesh.
The learned model $f_M$ is trained on data generated from the ground-truth simulator $f_S$, for which we use the OpenFOAM solver \cite{openfoam}.

\subsection{Model learning and optimization}

While each domain has a different state space structure and uses a different ground-truth simulator $f_S$, the learned simulators $f_M$ all share the same architecture (with identical hyperparameters in \twodfluids{} and \threedfluids{}, and only minor variations of the hyperparameters for \airfoil{} due to it being a steady state simulation; see \autoref{app:model}). 
The models are trained for next-step prediction on task-independent datasets (random perturbations of the design space for \airfoil{} and \threedfluids{}, and an open-source, qualitatively distinct dataset for \twodfluids{}), and are unrolled for up to 300 time steps during design optimization without further fine-tuning (see \autoref{sec:optimization} for details).

\section{Results}
\label{sec:results}

Our results show that learned simulators can be used to effectively optimize various designs despite significant domain shift and long rollout lengths.
The same underlying model architecture is used for each domain, highlighting the generality of learned simulators for design. 
Examples of designs found with our approach are available at: \url{https://sites.google.com/view/optimizing-designs}.
Performance is always evaluated using the ground-truth simulator $f_S$ (see \autoref{sec:optimization}).
Here we discuss these results, and compare the capabilities of gradient descent with learned simulators over classical simulators and sampling-based optimization techniques.

\subsection{Overall results}
\label{sec:overall-results}

We first asked whether a learned simulator combined with gradient descent (GD-M) could produce good-quality designs at all.
This approach might fail in various ways: accumulating model error, vanishing or exploding gradients \citep{bengio1994learning}, or domain shift \citep{hamrick2020role}.
However, as the following results show, GD-M produced high-quality designs across all three domains.

\autoref{fig:2dfluids_results}a shows qualitative results for \twodfluids{} (\autoref{sec:2d-fluid-design}), where our approach (GD-M) produces intuitive, functional designs to contain (\contain{}), transport  (\ramp{}), or funnel (\maze{}) the fluid to a target location.
On average, GD-M outperforms CEM-M by 16.1--118.9\%, indicating a substantial benefit of gradient-based optimization.
GD-M also outperforms CEM-S by 3.9--37.5\%, despite using a learned simulator rather than the ground-truth.
However, these design spaces are still relatively small (between 16 and 36 dimensions).
In \threedfluids{} (\autoref{sec:3d-fluid-design}), we substantially increase the dimensionality to a 625-dimensional landscape.
Here, GD-M produces robust designs, creating ridges to re-route water in particular directions or valleys to direct water into pools (\autoref{fig:3dfluids}a-c).
In comparison, CEM-M cannot solve any of these tasks, with performance 30--85$\times$ worse than GD-M.

In \airfoil{} (\autoref{fig:airfoil}b), GD-M recovers the characteristic S-curve shape for a low-drag airfoil under a small angle of attack, and matches the design obtained with \dafoam{}, an adjoint aerodynamics solver which computes close-to-optimal designs for this task.
Specifically, the design obtained with \dafoam{} yields a drag coefficient of 0.01902, while GD-M finds designs with drag between 0.01898--0.01919 depending on ensemble size (see \autoref{sec:ensembles}).
Importantly, \dafoam{}'s solver and optimizer are highly specialized for the particular task of airfoil design, while our approach is more general-purpose in that it requires only trajectory data for training and a generic gradient-based optimizer.

\begin{figure*}[!t]
\centering
\includegraphics[width=\textwidth]{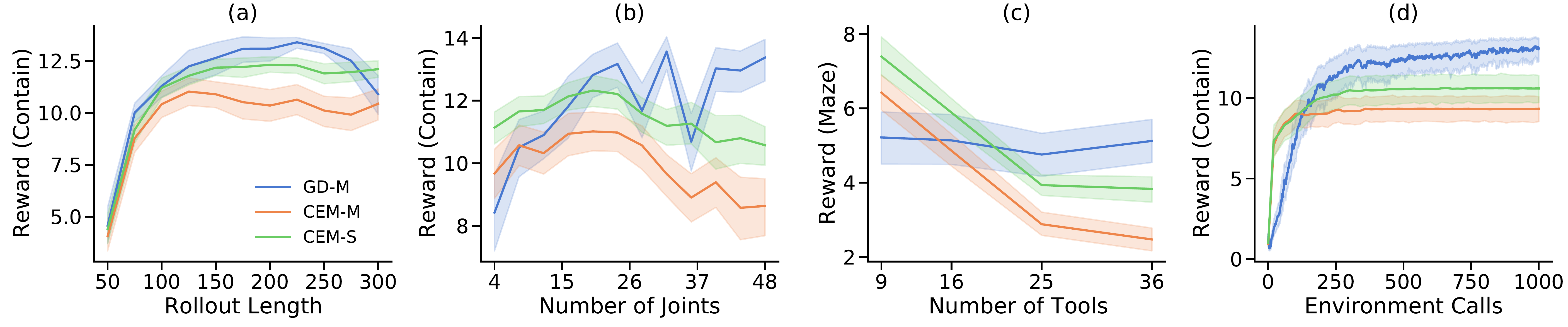}
\vspace{-1.6em}
\caption{
Ablation experiments on the \contain{} and \maze{} tasks. 
(a) Performance of all optimizers increase with rollout length; GD-M performance starts to deteriorate around step 225. (b) In \contain{}, CEM performance drops when increasing the number of joints above 24, while GD-M remains stable. (c) We observe a similar trend with the number of tools in \maze{}. (d) CEM often gets stuck in sub-optimal solutions early in optimization, while GD-M performance continues to increase.}
\vspace{-0.5em}
\label{fig:2dfluids_ablations}
\end{figure*}

\subsection{Model stability \& gradient quality}

We investigated accuracy over long timescales by measuring the effect of rollout length on design quality in \twodfluids{} (\autoref{fig:2dfluids_ablations}a and \ref{fig:hg_episode_sweep_demos}).
Longer rollouts can in principle allow for higher reward in this task as they give the fluid time to settle; however, with learned models, they can also be unstable due to error accumulation \citep{talvitie2014model,venkatraman2015improving}.
Nevertheless, we find that the learned simulator does not seem to be severely impacted by this problem.
Specifically, the quality of designs found by GD-M increases up to 225 steps (\autoref{fig:2dfluids_ablations}a), indicating that the learned simulator's accuracy and gradients remain stable for a surprisingly long time. 
Across the episode lengths evaluated, we find that GD-M outperforms not only CEM-M (by 18.1\% on average) but also CEM-S (by 4.4\% on average).
This indicates that the benefits of a having a learned model that supports better optimization techniques can outweigh the error incurred by long rollouts.

The strong performance of GD-M on longer rollout lengths is noteworthy.
Gradients tend to degrade when passed through chains of many model evaluations, and as a result, previous work generally only optimizes gradients in small action spaces over just a few time-steps \cite{li2018learning}.
We speculate that one reason for the success of GD-M is the addition of noise in training $f_M$, which promotes stability on the forward pass and may also force smoother gradients.

\subsection{Generalization}

Deep networks often struggle to generalize far from their training data \cite{geirhos2018generalisation}.
This poses a problem for design: to produce in-distribution training data, we would already need to know what good designs look like, thus defeating the aim of wanting to find \emph{new} designs.
However, we find that the GNN-based simulators studied here overcome this issue.
As noted in \autoref{sec:2d-fluid-design}, the learned simulator for \twodfluids{} was trained on a pre-existing, highly simplified dataset where only one to four straight line segments interact with a fluid 
(\autoref{app:domains}).
In contrast, the design tasks studied here involve highly articulated, curved obstacles (\contain{}, \ramp{}) or a larger number of obstacles (\maze{}); yet, GD-M still discovers effective designs without requiring any finetuning (\autoref{fig:2dfluids_results}).
We suspect this is because the model is trained to learn local collision rules, making it more robust to global distribution shift.

Using a learned simulator trained in a relatively simple environment has another unexpected advantage.
In rare cases, classical simulators suffer from degenerate behavior around certain edge cases.
For example, with the classical simulator for \twodfluids{}, particles can get stuck in between joint segments, especially when there are a large number of parts or joints (\autoref{fig:mpm_instability}). 
However, since the learned simulator was trained on simpler data where these effects are unobserved, it picks up only on the appropriate collision performance and not the unrealistic edge cases. 
Thus, the learned simulator produces more plausible rollouts than the classical simulator in these cases, and might therefore be a better candidate for producing designs that would transfer to the real world.

\subsection{Improving accuracy with ensembles}
\label{sec:ensembles}

In engineering tasks like \airfoil{}, simulators must be especially accurate, as small differences in the predicted pressure field can cause large errors in lift and drag coefficients.
While GD-M (without ensembling) can produce designs close to \dafoam{}'s, we notice a slightly rounder wing front (\autoref{fig:airfoil}b, bottom) causing a small increase in drag (0.01919 versus 0.01902 in \dafoam{}).

To further improve performance, we implemented an ensemble of 
learned simulators trained on separate splits of the training set.
Ensembles are a popular choice for training transition models for use in control \citep{chua2018deep}, as they can provide higher quality predictions and are more resistant to delusions---a particularly problematic issue for accuracy-sensitive domains such as airfoil design.
During optimization, we make predictions with all models in the ensemble, each trained on a different data split, and average the gradients.
As shown in \autoref{fig:airfoil}b-c, larger ensembles yield designs with significantly lower drag ($\beta = -5.3\times 10^{-5}$, $p = 0.0003$, where $\beta$ is a linear regression coefficient) and higher overall reward ($\beta = 5.6\times 10^{-5}$, $p = 0.0001$), and are able to produce designs very close to the solution found by \dafoam{}, with a drag coefficient of 0.01898 (size-5 ensemble).
Thus, with ensembles, we are able to achieve performant designs with a general-purpose learned simulator, indicating that we can use learned models for design optimization in spaces traditionally reserved for specialized solvers like \dafoam{}.

\subsection{Scalability to larger design spaces}
\label{sec:design-dimensionality}
In larger design spaces, sampling-based optimization procedures quickly become intractable, especially with relatively slow simulators.
We hypothesized that gradient descent with fast, learned simulators could overcome this issue, especially as the size of the design space is increased.
We therefore compared different optimizers on \twodfluids{} as a function of the dimensionality of the design space (the number of tool joints in \contain{} or the number of tools in \maze{}) and on the higher-dimensional \threedfluids{}.

For \contain{} (\autoref{fig:2dfluids_ablations}b), the performance of GD-M increases with the number of joints as increasingly fine grained solutions are made possible.
In contrast, for both CEM-M and CEM-S, design quality deteriorates with more joints as high quality solutions become harder to find with random sampling. In the highest dimensional \contain{} task with 48 tool joints, GD-M outperforms CEM-M by $154.9\%$ and CEM-S by $126.5\%$. 
When CEM does find solutions (\autoref{fig:2dfluids_results}b), they lack global coherence and appear more jagged than solutions found with GD-M.
Similarly, for \maze{} (\autoref{fig:2dfluids_ablations}c), the performance of GD-M is largely unaffected by the number of tools, while the performance of CEM-M and CEM-S both degrade as the design space grows. For the highest dimensional \maze{} problem with 36 joints, GD-M outperforms CEM-M by $207.4\%$ and CEM-S by $133.7\%$.

In the 625-dimensional \threedfluids{} task, CEM-M performs 30--85$\times$ worse than GD-M (\autoref{fig:3dfluids}) despite extensive hyperparameter tuning. This trend held across all tasks (\autoref{fig:accuracy_comparison}a), and even persisted when using fewer control points in the design space (\autoref{fig:design_space_3d}).
This is due not only to \threedfluids{}'s larger design space, but also because this problem requires a globally coherent solution: modifying small areas independently is unlikely to have much effect on the global movement of the fluid.

\subsection{Model speed and sample efficiency}

Learned simulators can provide large speedups over traditional simulators in certain domains by learning to compensate for coarser sub-stepping and making optimal use of hardware acceleration.
In \airfoil{}, although we use a very simple GD setup, our approach is able to find very similar designs as \dafoam{}'s specialized optimizer.
Moreover, our approach requires only 21s (single model) to 62s (size-5 ensemble) on a single A100 GPU, compared to 1021s for \dafoam{} run on an 8-core workstation, despite requiring $10\times$ more optimization steps. 

In \twodfluids{}, the ground-truth simulator runs at a similar speed to the learned model \citep[see][]{sanchez2020learning}, but is non-differentiable and therefore depends on more expensive gradient-free optimization techniques which require more function evaluations.
We use 20--40 function evaluations per optimization step of CEM, compared to a single evaluation with GD (which is about $3\times$ more costly, due to the gradient computation). Thus, GD with a differentiable learned model can be much more efficient than using the ground truth simulator with a sampling-based method.
\section{Discussion}

We used state-of-the-art learned, differentiable physics simulators with gradient-based optimization to solve challenging inverse design problems.
Across three domains and seven tasks, which involved designing landscapes and tools to control water flows or optimizing the shape of an airfoil, we demonstrated that gradient descent with pre-trained simulators can discover high-quality designs that match or exceed the quality of those found using alternative methods.
This approach succeeds in a variety of interesting and surprising ways:
it permits gradient backpropagation through complex physical trajectories for hundreds of steps; scales to tasks with large design and state spaces (100s and 1000s of dimensions, respectively); and successfully generates designs which require the learned simulator to generalize far beyond its training data.
In the classic aerodynamics problem of airfoil shape optimization, our approach produces a design comparable to that of a specialized solver using only simple, general-purpose strategies like model ensembling.

While our results have exciting implications for inverse design, they also open up possibilities for explaining everyday human behavior like tool invention---a longstanding puzzle in cognitive science \citep{allen2019tools,osiurak2016,Shumaker2011}. 
With general-purpose learned simulators, we have the potential not just to create highly specialized tools in engineering domains, but also to model everyday tool creation, such as creating a hook from a pipe cleaner, building a blanket fort, or folding a paper boat.

Our approach has limitations that should be visited in future work.
Gradient descent is inappropriate for design spaces with regions of zero gradients, such as in \twodfluids{} tasks where the fluid may not always make contact with the tool (\autoref{fig:mpm_zero_grad}).
Many interesting design tasks also have variably-sized or combinatorial design spaces that cannot easily be optimized with gradient descent, such as computer-aided design (CAD) approaches to 3D modeling.
An exciting future direction will be to integrate general-purpose learned simulators with hybrid optimization techniques such as those used in material science and robotics \citep{chen2020generative,toussaint2018differentiable}.
As learned simulators continue to improve, we could also use them to do even broader cross-domain, multi-physics design.
While challenges remain, our results represent a promising step towards faster and more general-purpose inverse design.

\bibliography{inverse_design}
\bibliographystyle{icml2022}

\newpage
\appendix
\onecolumn

\appendix

\renewcommand\thefigure{A.\arabic{figure}}
\renewcommand\thetable{A.\arabic{table}}
\setcounter{figure}{0} 
\setcounter{table}{0} 

\section{Optimizer hyperparameters}
\label{app:hyperparameters}
Optimization hyperparameters were chosen to reflect good performance for each optimizer in each domain. 
We therefore performed sweeps for major hyperparameters of each optimizer for each domain, with those used for experiments in the paper shown in the table below.

\begin{figure}[H]
    \centering
    \includegraphics[width=0.7\textwidth]{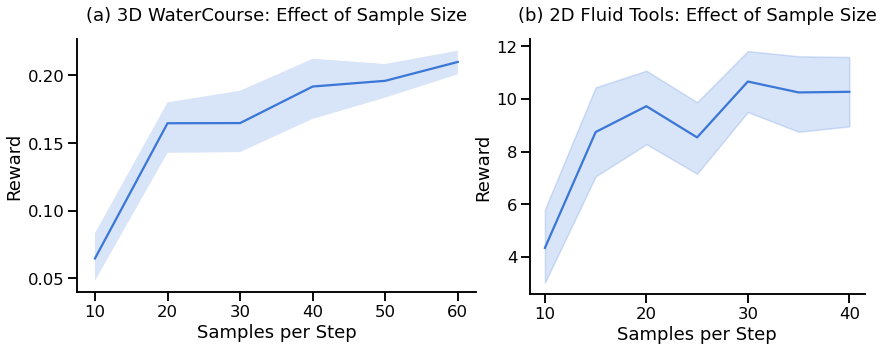}
    \caption{For CEM, increasing population size, while more computationally expensive, can lead to improvements in performance. (a) In the \threedfluids{} domain, CEM benefits from large sample sizes, although returns are diminishing for sizes beyond 40 (\direction{} task, 36 design parameters). (b) In the \twodfluids{} domain, CEM benefits from larger sample sizes, although returns are diminishing for sizes beyond 20 (\contain{} task, 40 design parameters).}
    \label{fig:cem_samples}
\end{figure}

CEM maintains a population of samples and uses these to estimate the mean $\mu$ and standard deviation $\sigma$ of a Gaussian distribution over design parameters. To optimize $\mu$ and $\sigma$, it takes the top performing fraction, deemed the ``elite portion,'' from the current step. The initial standard deviation of this distribution is given by ``Initial $\sigma$'', and the initial mean is set to 0.
We found that for CEM, the population sample size had a significant effect on overall optimization quality (\autoref{fig:cem_samples}). Due to computational considerations, we picked the smallest value for this hyperparameter that performed within 1 standard deviation of the optimal sample size. Both the elite portion and initial $\sigma$ parameters were chosen as the best performing values on a set of held-out random tasks for each domain.

For GD, we only performed a hyperparameter sweep over the learning rate, which was the only parameter to significantly affect performance. For the \twodfluids{} tasks, we introduced gradient clipping to eliminate the effect of rare gradient spikes over the course of optimization. However, not using gradient clipping still produced qualitatively and quantitatively similar results. For additional Adam parameters, we used the default values for the exponential decay rates that track the first and second moment of past gradients of $b_1=0.9$ and $b_2=0.999$ \cite{optax}.

\begin{center}
\begin{tabular}{ r | c c c c c c}
\toprule
& \multicolumn{3}{c}{\twodfluids{}} &   \multicolumn{2}{c}{\threedfluids{}}         &   \airfoil{}  \\
GD	                &	\contain{}	&	\ramp{}	&	\maze{}    &   \direction{}   &   \emph{Pools}	&               \\
\midrule							
Learning rate	    &	0.005	&	0.005	&	0.01    &   0.01        &   0.01    &   0.01     \\
Momentum term $b_1$ &	0.9 	&	0.9 	&	0.9     &   0.9	        &   0.9     &   0.9     \\
Momentum term $b_2$ &	0.999 	&	0.999 	&	0.999   &   0.999	    &   0.999   &   0.999     \\
Gradient clip   	&	10  	&	10  	&	10      &   ---	        &  ---      &   ---     \\
                    &           &           &           &               &           &        \\
CEM                 &           &           &           &               &           &        \\
\midrule							
Sampling size   	&	20	    &	20	    &  	20      &   40	        &   40      &   ---     \\
Elite portion	    &	0.1	    &	0.1	    &	0.1     &   0.1	        &   0.1     &   ---     \\
Initial $\mu$       &	0	    &	0	    &	0       &   0	        &   0       &   ---     \\
Initial $\sigma$    &	0.5	    &	0.5	    &	1.5     &   0.1	        &   0.1     &   ---     \\
Evolution smoothing	&	0.1	    &	0.1	    &	0.1     &   0.1	        &   0.1     &   ---     \\
\midrule							
Optimization steps   	&	1000	    &	1000	    &  	1000      &   200	    &   200 &   200     \\
\bottomrule
\end{tabular}
\label{tab:optimizer_hypers}
\end{center}

\section{Model architecture and training}
\label{app:model}
For each task domain, we train a GNN for next-step prediction of the system state. 
For the domains considered in this paper, we unify the approaches of GNS \citep{sanchez2020learning} and \mgn{} \citep{pfaff2021learning}:
In the \airfoil{} domain, we encode/decode mesh nodes and mesh edges as a graph as described in the aerodynamics examples of \mgn{}, while for particle-based fluids, edges are generated based on proximity as in GNS. In the case of \threedfluids{}, both particles (fluid) and a mesh (the designed obstacle) are present; hence, edges are generated based on proximity (for fluid-fluid and fluid-obstacle interaction) or from the landscape mesh. As the landscape does not have any internal dynamics, we did not find it necessary to distinguish between world- and mesh edges, and use a single edge type.

Once encoded as a graph, the core model and training procedure is largely identical between GNS and \mgn{}, and we refer to the above papers for full details on architecture and model training. 
Briefly, we use an Encode-Process-Decode GNN with 10 processor blocks. All edge and node functions are 2-layer MLPs of width 128, with ReLu activation and LayerNorm after each MLP block. The model is trained with Adam and a mini-batch size of 2, with training noise, for up to 10M steps. We implemented this model in JAX \citep{jax2018github}.
In addition to the different encoding procedures for mesh vs. particle systems, the parameters for training noise and connectivity radius have to be set per-domain, to account for differences in particle size/mesh spacing. These details are described in \autoref{app:domains}.

\paragraph{Gradient computation}
\label{app:gradients}
In our experiments, we pass gradients through long model rollouts of up to 300 steps. As it is prohibitive to store all forward activations for the backwards pass, we use gradient checkpointing \citep{chen2016training} to store activations only at the beginning of each step of the trajectory during the forward pass, and recompute the intermediate activations for each step as needed when the backwards pass walks the trajectory in reverse. Gradient calculation using this method has roughly 3 times the time cost of a pure forward simulation: forward dynamics have to be computed twice for each step, in addition to the computation of the backwards pass itself.

\section{Task domains}
\label{app:domains}

\subsection{\twodfluids{}}
\label{app:2dfluids}
Tasks in \twodfluids{} are procedurally generated from templates specified in \autoref{tab:2dfluids_tasks}.
The simulation domain is a 2D box, with the lower left corner specified as $[0, 0]$, and upper right corner specified as $[1,1]$. Fluid particles are initialized as a box of size \emph{Initial fluid box} with bounding boxes given in format $[x_\text{min}, y_\text{min}, x_\text{max}, y_\text{max}]$. Certain task parameters were varied for ablation experiments in \autoref{fig:2dfluids_ablations} (rollout length, \# joints (\contain{}), \# tools (\maze{})); \autoref{tab:2dfluids_tasks} contains default values used unless otherwise specified.

\begin{table}[t]
\begin{center}
\begin{tabular}{ r | c c c }
\toprule
	&	Contain	&	Ramp	&	Maze (nxn)	\\
\midrule
Environment size	&	1x1	&	1x1	&	1x1	\\
Rollout length	    &	150	&	150	&	150	\\
\\
Initial fluid box	&	[0.2, 0.5, 0.3, 0.6]	&	[0.2, 0.5, 0.3, 0.6]	&	[0.2, 0.75, 0.8, 0.8]	\\
\\
Reward sampling box 	&	[0.4, 0.1, 0.6, 0.3]	&	[0.8, 0, 1, 0.2]	&	[0.1, 0.1, 0.9, 0.2]	\\
Reward $\sigma$	        &	0.1	                    &	0.1             	&	0.1	                    \\
\\
Design parameter	&	joint angles    &	joint angles	&	rotation	\\
\# tools	        &	1           	&	1	            &	$n^2$   	\\
\# joint angles	    &	16	            &	16	            &	1	        \\
\\
Tool position (left)	    &   [0.15, 0.35]	&	[0.15, 0.35]	&	---	\\
Tool domain box (3x3)      	&	---	            &	---	            &	[0.14, 0.3, 0.65, 0.6]	\\
Tool domain box (4x4)      	&	---	            &	---	            &	[0.14, 0.3, 0.71, 0.6]	\\
Tool domain box (5x5)      	&	---	            &	---	            &	[0.14, 0.3, 0.75, 0.6]	\\
Tool domain box (6x6)      	&	---	            &	---	            &	[0.14, 0.25, 0.77, 0.65]	\\
\\
Tool Length	        &	0.8	&	0.8	&	---    	\\
Tool length (3x3)	&	---	&	---	&	0.72	\\
Tool length (4x4)	&	---	&	---	&	0.64	\\
Tool length (5x5)	&	---	&	---	&	0.65	\\
Tool length (6x6)	&	--- &	---	&	0.63	\\
\bottomrule
\end{tabular}
\end{center}
\caption{Task Parameters for \twodfluids{} tasks. Boxes are described as $[x_\text{min}, y_\text{min}, x_\text{max}, y_\text{max}]$.}
\label{tab:2dfluids_tasks}
\end{table}

\begin{figure}[t!]
\centering
\includegraphics[width=0.8\textwidth]{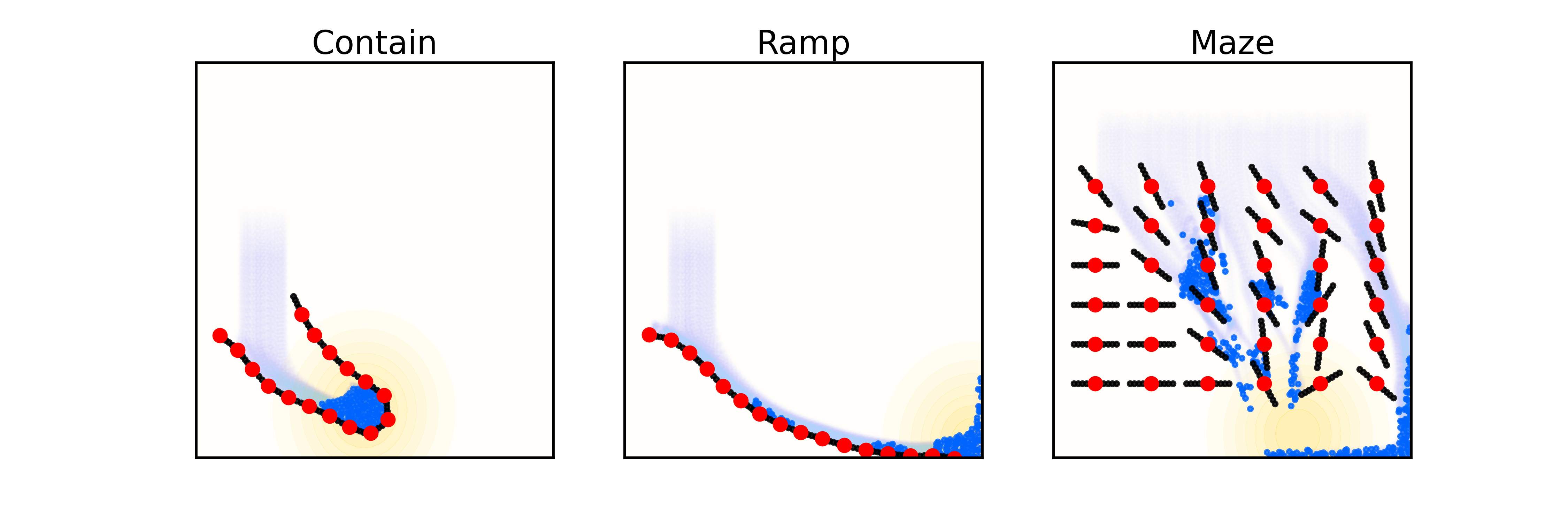}
\caption{Visualization of the design space parameterization for the \twodfluids{} task. Each red dot corresponds to the anchor points (\contain{} and \ramp{})  and center of rotation (\maze{}) being optimized.}
\label{fig:hg_designspace_appendix}
\end{figure}

\paragraph{Design space}
A ``tool`` in this task domain is a 2D curve composed of several line segments connected by joints. For a large number of joints, a tool can thus approximate a smooth curve (\autoref{fig:hg_designspace_appendix}).
Each task's design space consists of the relative joint angles controlling the tool's shape. We consider tasks with a single, multi-segment tool (\contain{}, \ramp{}) and a task with multiple, single-segment tools (\maze{}).
For each tool, relative angles are calculated by moving from the anchor point on the left, along the tool segments to right, such that $\text{angle}_i = \text{angle}_{i-1} + {\phi_\mathrm{joints}}_i$ for the $i^\text{th}$ joint from the anchor.
We also experimented with two additional design space parameterizations: (1) jointly optimizing the joint angles and a global position offset $[x, y]$ for each tool, and (2) changing the parameterization of angles to be absolute (such that $\text{angle}_i = {\phi_\mathrm{joints}}_i$ directly). 
We discuss the effects of these alternate parameterizations in \autoref{app:design-param-results}.

\paragraph{Simulation and objective}
Both fluids and tools are represented as particles with different types, and simulated with the learned model for 150 steps (with the exception of the ablation experiment on rollout length). Scenes consist of  $N=100 \ldots 1000$ fluid particles.
For ground-truth evaluation of the designs, we simulate particle dynamics with an MPM solver~\cite{hu2018moving}. 
Task reward is calculated using the Gaussian likelihood of the final particle positions after rollout ($\mathbf{u_v}$ from $\tilde{G}^{t_K}$). That is, for a task with reward parameterized with mean $\mu$ and spherical covariance $\sigma$ ($\theta_R = [\mu, \sigma]$), the reward is calculated as 
\begin{equation*}
    f_R := \mathrm{mean}_v]\, \mathcal{N}(\mathbf{u_v}; \mu, \sigma)\,.
\end{equation*}

\paragraph{\contain{}}
For this task, the center of the goal region $\mu$ is sampled uniformly from a rectangular reward region in the lower-middle section of the $1\times1$ simulation domain  ($[0.4, 0.6]\times[0.2, 0.4]$). 
A tool protruding to the right is initially placed below the fluid rectangle.
By optimizing a single tool's relative joint angles, successful solutions must ``contain'' the fluid in the region by creating a cup or spoon. 

\paragraph{\ramp{}}
The fluid and tool are initialized as in \contain{}, and $\mu$ is sampled from a region lower and further to the right than in \contain{} ($[0.8, 1]\times[0, 0.2]$). 
By again optimizing a single tool's relative joint angles, successful solutions will create a ``ramp'' from the initial fluid position to the goal location in the bottom right.

\paragraph{\maze{}}
The goal is sampled from a long region near the bottom of the domain ($[0.1, 0.9]\times[0.1, 0.2]$).
By optimizing the rotation angles of a grid of rigid, linear tools, successful solutions will create a directed path from the top of the screen to the goal location at the bottom.

\paragraph{Model training}
We trained the learned simulator on the \textsc{WaterRamps} datasets released by \citet{sanchez2020learning}. This dataset consists of 1000 trajectories featuring a single large block of water falling on one to four randomized straight line segments (see image below for examples). Model architecture and hyperparameters are described in \autoref{app:model}, with a training noise scale of $6.7\,10^{-4}$ and connectivity radius of $0.015$.

\begin{figure}[H]
\begin{center}
\includegraphics[width=0.7\textwidth]{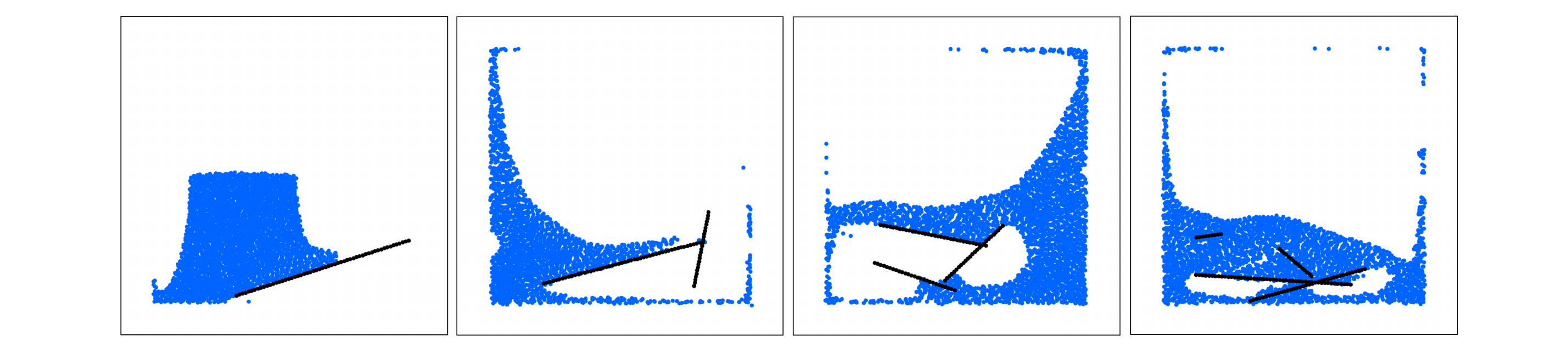}
\caption{Four examples of trajectories from the \textsc{WaterRamps} dataset released by \citep{sanchez2020learning} used as training data for the supervised prediction model.}
\end{center}
\end{figure}

\subsection{\threedfluids{}}
\label{app:3dfluids}

\paragraph{Design space}
This domain has a design space $\phi_\mathrm{map}$ of 625 parameters, which determine the y coordinate offset to nodes of a $25 \times 25$ square mesh centered at $\mathbf{c}=(0.5, 0.5, 0.5)$ in the simulation domain.
While we could directly mapping the parameters to coordinates, we use the design function $y_i = \gamma_H \mathrm{tanh}(\phi_{\mathrm{map}_i})$ ($\gamma_H = 0.3$ for all tasks) to prevent trivial task solutions (i.e. obstacles which touch the floor). 

\paragraph{Simulation}
The simulation consists of an inflow pipe located at (-0.5, 1.0, 0.5) above the landscape which continually emits a stream of liquid, represented as particles. These particles are then redirected by the designed landscape, and finally removed once they hit the floor at $y=0$. 
In our experiments we observed up to 2084 particles present in the scene at one time. We unroll the learned simulation model for trajectories of 50 time steps, and store the final particle positions $\mathbf{u_v}$, as well as the positions of removed particles that touched the floor at any point $\mathbf{u_v^D}$ to be passed to the reward function.
Ground truth simulations for evaluation are performed by running the same setup with an SPH solver. We note that SPH requires very small simulation time steps, and performs $\approx 10^4$ internal steps for a trajectory of the same length.

\paragraph{\direction{}}

In this task, we want to align the water stream with a given direction vector $\mathbf{d}$. We can formalize this using the reward function 
\begin{equation*}
f_{R_\mathrm{dir}} := \mathrm{mean}_v \left( (\mathbf{u_v} - \mathbf{c})\cdot \mathbf{d}\right) - \mathrm{std}_v \left( (\mathbf{u_v} - \mathbf{c})\cdot\mathbf{d_\perp}\right) - \gamma_R\, \mathrm{mean}(\nabla \phi_\mathrm{map})
\end{equation*} 
where $\mathbf{d}_\perp$ is orthogonal to $\mathbf{d}$. The first term aligns the direction of the particle relative to the domain center, and the second term concentrates the stream. The last term is a smoothness regularizer on the design landscape, which prefers smooth solutions ($\gamma_R=300$ for both tasks). Absolute reward numbers for this task can be positive or negative, hence we report the normalized reward $f_{R_\mathrm{dir}} - f_{R_\mathrm{dir}}^\mathrm{initial}$, i.e. an unchanged initial design corresponds to a zero reward, to make the scores easier to interpret.

Rewards can be in 8 different directions, spaced between $0$ and $180\deg$. We collapse across directions for reporting reward means and confidence intervals for each optimizer.

\paragraph{\twopools{} and \threepools{}}
In these tasks, we define two and three pools, respectively, with center $\mathbf{\mu_p}$ on the floor. For each particle which has hit the floor, we assign it to its closest pool $\mathbf{\hat{\mu_p}}$, and define the reward as the Gaussian probability under $\mathbf{\hat{\mu_p}}$, i.e.
\begin{equation*}
f_{R_\mathrm{pools}} := \mathrm{mean}_v (\mathcal{N}(\mathbf{u_v^D};\mathbf{\hat{\mu_p}, \sigma})) - \gamma_R\,\mathrm{mean}(\nabla \phi_\mathrm{map})
\end{equation*} with $\sigma=0.4$ and a regularization term as above.

To showcase different ways of splitting the water stream, we consider one positioning of the pools for the two pool case, and two for the three pool case. In the two pool case, pools are placed at $[1.49, -0.35]$ and $[1.49, 1.35]$. In the three pool case, pools are placed either at $[1.6, -0.45]$, $[1.85, 0.5]$, and $[1.6, 1.45]$, or at $[0.5, -0.5]$, $[1.7, 0.5]$, and $[0.55, 1.5]$. These were selected to ensure the task was solveable -- pools directly beneath the landscape, or too far away from the landscape, would not be reachable even with dramatically warped surfaces.

\paragraph{Model training}
We trained a model on next-step prediction of particle positions, on a dataset of 1000 trajectories of water particles interacting with a randomized obstacle plane (random rotations and sine-wave deformations of the planar obstacle surface). The data was generated using the SPH simulator SPlisHSPlasH \citep{bender2015divergence}.
The noise scale is set to $0.003$ and a connectivity radius of $0.01$ to account for the different particle radius of the 3D SPH simulation compared to 2D MPM. All other architectural and hyperparameters are as described in \autoref{app:model}.

\subsection{\airfoil{}}
\label{app:airfoil}
The airfoil optimization task is modeled similarly to the NACA0012 aerodynamic shape optimization configuration for incompressible flow for the DAFoam solver (see details \href{https://dafoam.readthedocs.io/en/latest/Tutorial_Aerodynamics_NACA0012_Incompressible.html}{here}), to make it easier to compare design solutions to this solver.

\paragraph{Design space}
The design space consists of the $y$-coordinate of 10 control points (see \autoref{fig:airfoil}a). Moving these control points deforms both the airfoil, and the simulation mesh surrounding it. The airfoil shape is deformed using B-spline interpolation as described by \citet{ffd}, and the mesh is deformed using IDWarp~\citep{Secco2021}.
We thus define a design function $G^{t0} = f_D(\phi_\mathrm{ctrl}, G_{\alpha})$ which takes an initial, undeformed airfoil mesh (we use the standard NACA0012 airfoil), encoded as a graph $G_{\alpha}$, as well as the control point position $\phi_\mathrm{ctrl}$ as input. It returns the graph of the deformed airfoil mesh $G^{t_0}$ to be passed to the simulator.
We note that the coefficients for spline interpolation and mesh warping can be precomputed for a given initial mesh, making it easy to define a differentiable function to use for design optimization.

\begin{figure}[t!]
\centering
\includegraphics[width=0.6\textwidth]{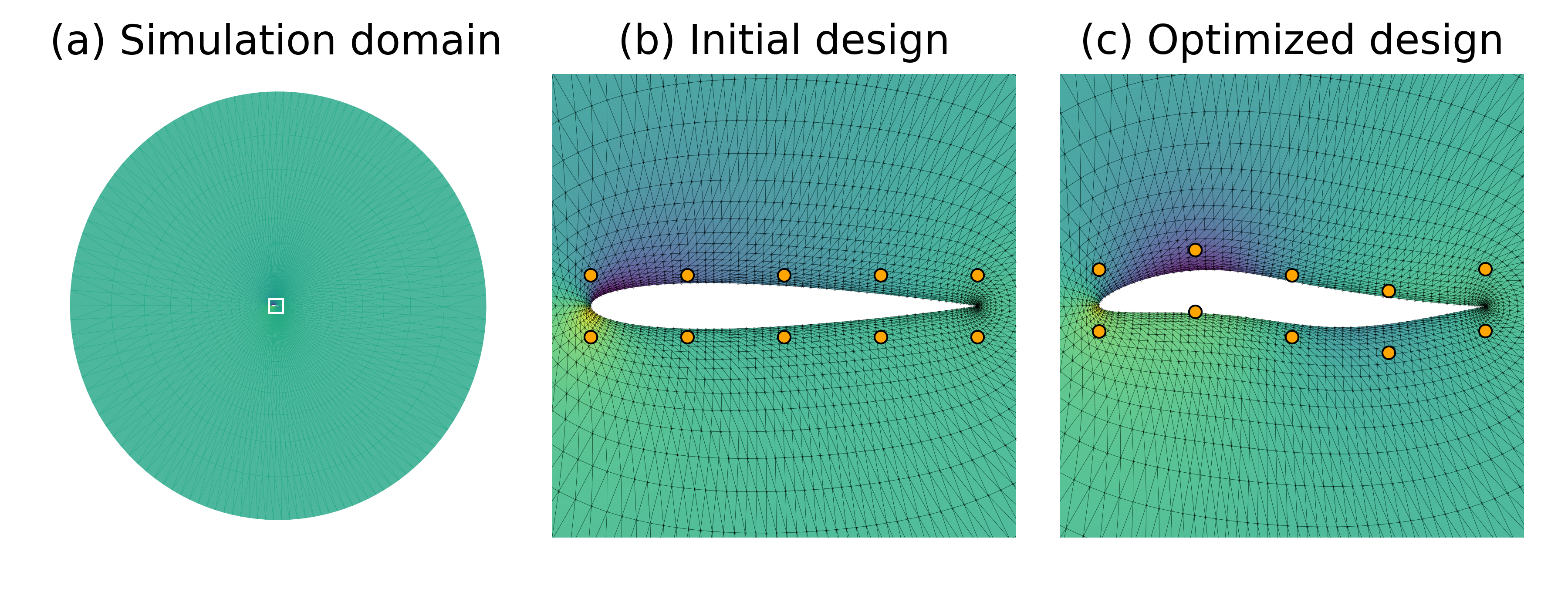}
\caption{(a) Aerodynamics are computed on a large 4158 node mesh centered around the airfoil, with the closeup regions around the airfoil in (b, c) marked as a white square in the center. (b, c) Pressure predictions and control points (orange) for the initial and final optimized wing design (Ensemble-5 model).}
\label{fig:airfoil_appendix}
\end{figure}

\paragraph{Simulation}
Given the initial mesh, as well as simulation parameters, the simulator or learned model predict the steady-state incompressible airflow around the wing, sampled on each of the 4158 nodes on the simulation mesh. The entire simulation domain and an example prediction of the pressure field are shown in \autoref{fig:airfoil_appendix}a,b.
For drag minimization we require predictions of the pressure field $p$, as well as the effective Reynolds stress $\rho_\mathrm{eff}$ at each mesh node, i.e. $\mathbf{q}_v=(p,\rho_{eff})$. Unlike the other domains in this paper, this is a single-step prediction task, and model rollouts are of length one.
For this task, we consider an inflow speed of 0.1 mach, under an $5.1^{\circ}$ angle of attack.

\paragraph{Task objective}
The task reward is defined as $f_R := -C_D - \gamma_L ||C_L - C_{L0}||^2 - \gamma_A\, a(\phi_\mathrm{ctrl})$, i.e. we minimize the drag coefficient $C_D$ under soft constraints of unchanged lift $C_L$ and a wing area $a$ of 1-3 times the initial area. We use $\gamma_L = 10, \gamma_A = 1$, and a tanh nonlinearity to enforce the volume inequality. Lift and drag can be computed from the simulation output $p, \rho_{eff}$ by integration around the airfoil, see e.g. \citep{ladson1988effects}. We report the normalized reward $f_R - f_R^\mathrm{initial}$ such that the initial, undeformed wing design corresponds to a zero reward.

\paragraph{Model training}
We trained a model to predict $p, \rho_{eff}$ on a dataset of 10000 randomized airfoil meshes, simulated with OpenFoam \citep{openfoam}. For training ensemble models, this dataset is split into 5 non-overlapping blocks, and a separate model is trained on each section. Since this is a steady-state prediction task, information needs to propagate further at each model evaluation. We therefore use twice-repeated processor blocks with shared parameters, i.e. the model performs 20 message passing steps, with 10 blocks of learnable parameters. We found that this increases accuracy in the one-step setup by being able to pass messages further across the mesh.
Training noise is often cited for stability over long rollouts, but even in this one-step setting, training noise and data variation can be useful. To increase robustness to unseen wing configurations, we varied the grid resolution between 1000-10000 nodes for each sample in the training set, and added training noise to the input mesh coordinates. We use a normal noise distribution with the scale of $1\%$ of the average edge lengths surrounding the node noise is applied to.
All other aspects of model architecture and training procedure are as described in \autoref{app:model}.

\section{Further results}

\subsection{Model accuracy}
\label{app:model-accuracy}

\begin{figure}[H]
\centering
\includegraphics[width=0.96\textwidth]{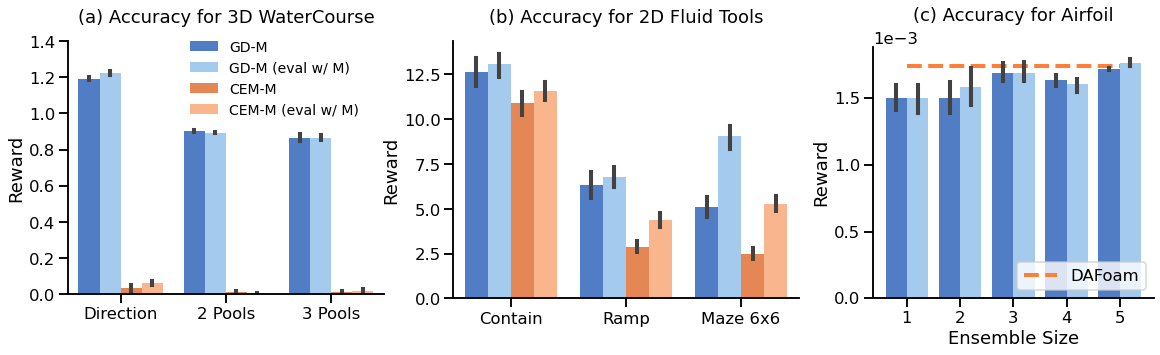}
\vspace{-1em}
\caption{(a) For the \threedfluids{} domain, reward predicted by the learned model (GD-M eval w/ M, CEM-M eval w/ M) is very close to the ground-truth simulator evaluation (GD-M, CEM-M) for all tasks. (b) This is also true for the \twodfluids{} domain, though the reward is slightly overestimated when using the model. This effect is amplified in \maze{}, where the ground-truth dynamics sometimes struggles to correctly simulate ``sticky'' bottlenecks (see \autoref{fig:mpm_instability}). (c) Model predictions of drag (GD-M eval w/ M) are relatively close to the ground-truth simulator evaluation (GD-M), particularly for larger ensemble sizes.}
\label{fig:accuracy_comparison}
\end{figure}

In order for a learned simulator to be useful for design, it must be sufficiently accurate in the forward direction. We study this question directly for each of the domains (\threedfluids{}, \twodfluids{} and \airfoil{}) by examining the magnitude of the error between the model predictions of reward for the discovered designs, and the ground truth reward for those designs. The results for each domain are shown in \autoref{fig:accuracy_comparison}.

Broadly, the learned model very successfully mimics the ground truth simulator in reward prediction across all three domains. The accuracy for \airfoil{} is within a single standard deviation across all ensemble sizes, while the predictions in \threedfluids{} match very closely for both the high performing designs (GD-M) and low performing designs (CEM-M).

However, we do notice some discrepancies in the predicted and ground truth reward for the \twodfluids{} domain, particularly the \maze{} task. As mentioned in the main text, the ground truth solver sometimes produces unrealistic rollouts for this domain (see \autoref{fig:mpm_instability}), with fluid particles becoming stuck between the different tools. Despite this issue, we find that the model is sufficiently similar to the ground truth to produce designs that still achieve high reward overall.

\subsection{Effects of design parameterization}
\label{app:design-param-results}
In this section, we study how different parameterization choices for the design space affect both gradient-based and sampling-based optimizers. 

\begin{figure}[H]
\centering
\hspace{-4mm}\includegraphics[width=0.8\textwidth]{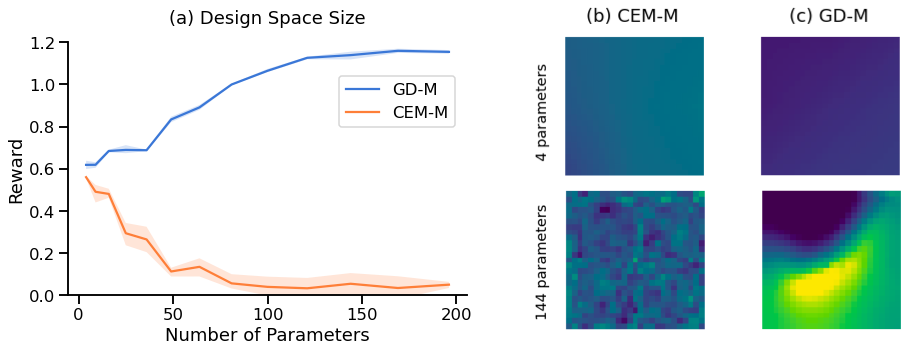}
\caption{Performance on the \threedfluids{} \direction{} task with variable design space resolution: GD performs well for large design spaces, while CEM performance quickly drops with increased number of design parameters. (b) Examples of CEM designs at two different design space resolutions. (C) Examples of GD designs at two different design space resolutions.}
\label{fig:design_space_3d}
\end{figure}

First, in the \threedfluids{} domain, we investigate a parameterization of the design space that uses interpolation to minimize the number of control points on the 2D heightfield (\autoref{fig:design_space_3d}). Control points are placed evenly across the grid, and bi-linearly interpolated onto the $25 \times 25$ mesh. We vary the number of control points from $2\times2$ up to $14\times14$, and find that while CEM-M performs similarly to GD-M when very few control points are allowed, its performance quickly drops as more control points are added.

Second, in the \twodfluids{} domain, we investigate what happens when we change the design space to use \emph{absolute} joint angles rather than relative ones. When using relative joint angles, changes to joints near the tool's pivot (left side) affect the global properties of the tool. We hypothesized that this could be selectively benefiting the sampling-based approaches, as this makes the effective design space much lower dimensional. We therefore change the design space to be absolute, with a tool's joint angles calculated directly: $angle_i = \phi_i$. 

\begin{figure}[H]
\centering
\includegraphics[width=0.8\textwidth]{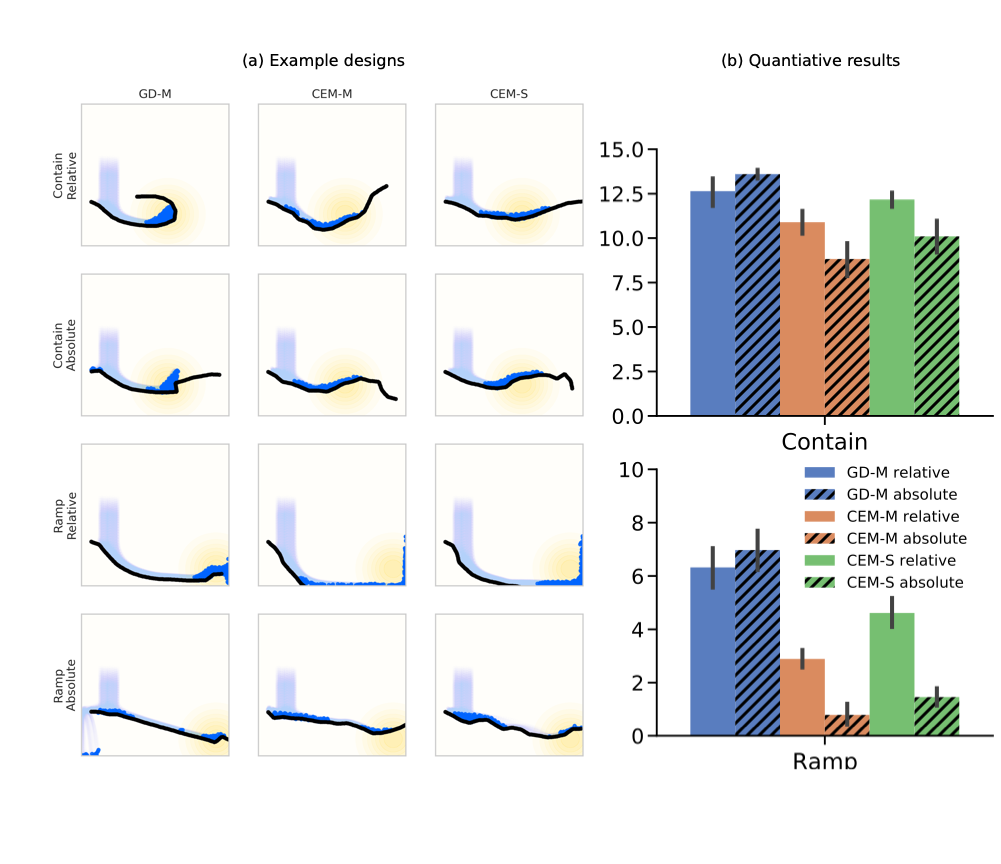}
\vspace{-2em} 
\caption{\textbf{(a)} Example solutions for each optimizer across the \contain{} and \ramp{} tasks when optimizing over \textit{Relative} vs \textit{Absolute} angles. \textbf{(b)} Mean reward with $95\%$ confidence intervals obtained by each optimizer across the \contain{} and \ramp{} tasks when optimizing over \textit{Relative} vs \textit{Absolute} angles.}
\label{fig:absolute_vs_relative}
\end{figure}

As hypothesized, this change does dramatically decrease the performance of the sampling-based technique (see \autoref{fig:absolute_vs_relative}). Perhaps more surprisingly, the gradient-based optimizer is almost completely unaffected by this reparameterization. While the qualitative solutions it finds differ (with tools now containing ``kinks'' to prevent the motion of the fluid rather than curves, \autoref{fig:absolute_vs_relative} top), the overall reward achieved is similar.

\subsubsection{Failure modes of gradient descent}
\label{app:twodfluids-zero-gradients}

Other parameterizations of the design space can badly affect the performance of gradient-based optimizers. In particular, gradient-based optimizers suffer when there are regions of zero gradients. In the \airfoil{} and \threedfluids{} domains, this is not normally a problem, as the design always interacts with the physical system on which reward is being measured. But in the \twodfluids{} domain, we can manipulate this.

\begin{figure}[H]
\centering
\includegraphics[width=0.74\textwidth]{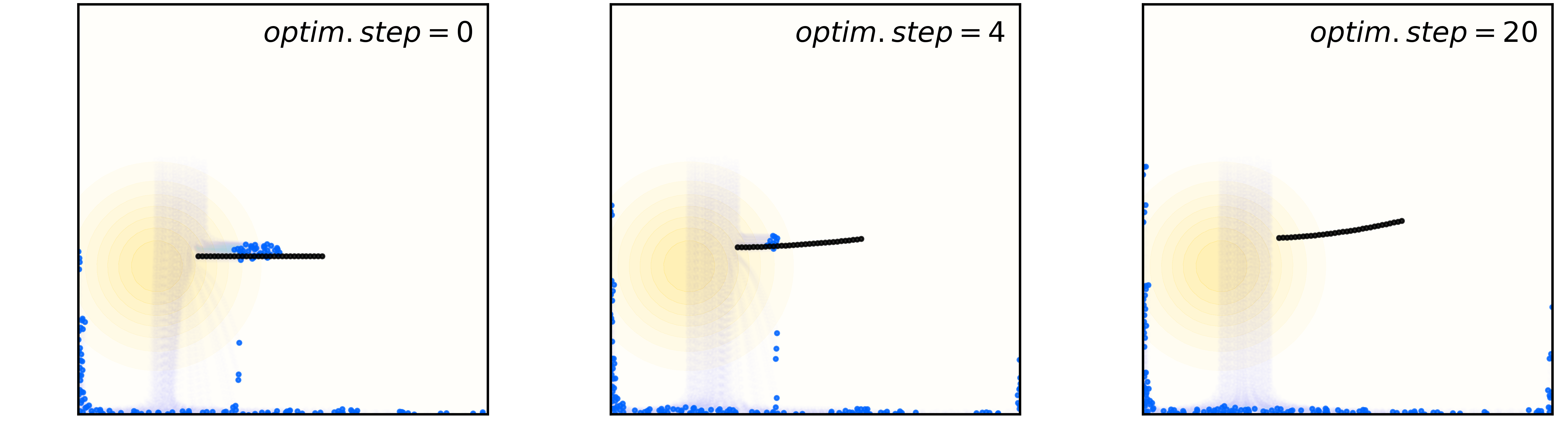}
\includegraphics[width=0.24\textwidth]{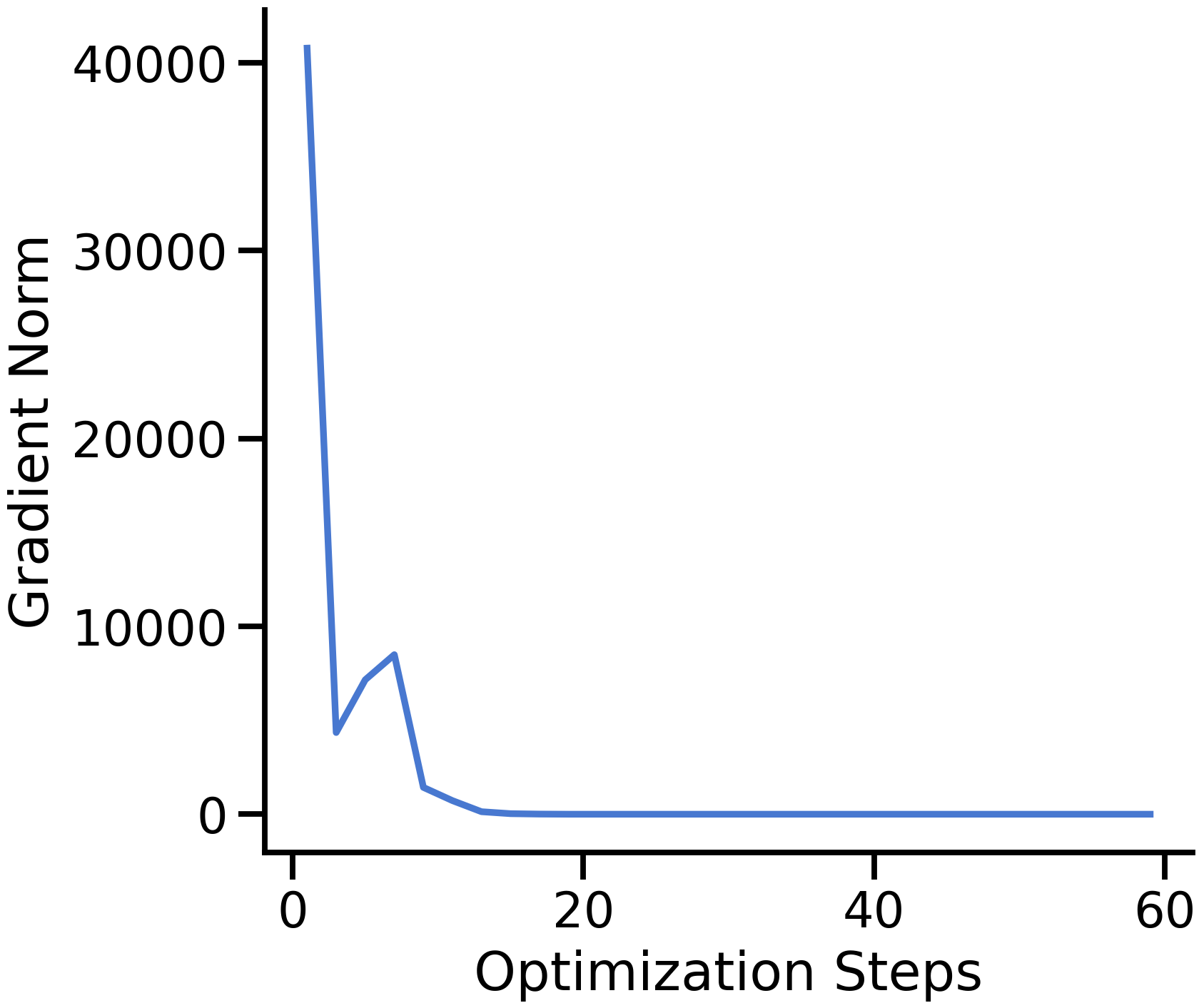}
\caption{Failure mode of the GD optimizer: in some instances where the translation of the tool is included in the design, the tool may end up outside of the scope of the fluid. In this cases, the optimization can no longer recover as it will get zero gradients from there on.}
\label{fig:mpm_zero_grad}
\end{figure}

In particular, for this experiment we changed the design space for \twodfluids{} to include a global position offset $[x,y]$ for the tool. Making this simple change often has no effect on the discovered designs, but occasionally the gradient-based optimization procedure can move the tool such that it no longer interacts with the fluid (\autoref{fig:mpm_zero_grad}). Once the tool has been moved out of the range of the fluid, there is no longer any way to affect the reward, and therefore there is no gradient signal to recover. To overcome this problem, future work would need to consider more sophisticated hybrid optimization techniques \cite{toussaint2018differentiable}.

\subsection{Failure modes of the MPM solver}
\label{app:twodfluids-mpm-instability}

\begin{figure}[H]
\centering
\includegraphics[width=0.5\textwidth]{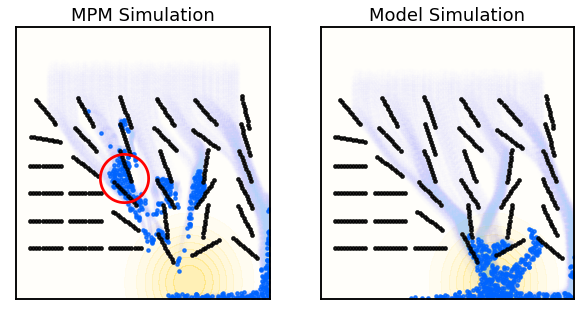}
\caption{Left: MPM simulation \citep{hu2018moving} of a problem with many separate solid objects. As highlighted in the red circle, the MPM solver struggles with water movement between obstacles, often creating artificially sticky bottlenecks. Right: Learned model rollout for the same setup. The model rollout looks significantly more plausible, without any ``stickiness'' artifacts. Please see \url{https://sites.google.com/view/optimizing-designs} for videos demonstrating this effect clearly.}
\label{fig:mpm_instability}
\end{figure}

One of the advantages of using a learned simulator over a classic simulator is learned simulators can be trained in regions of the state and action space that are known to exhibit regularized, smooth behaviors. For example, as mentioned in \autoref{app:model-accuracy}, the MPM solver \cite{hu2018moving} we use for evaluation in \twodfluids{} shows surprising irregularities with ``sticking'' behavior when there are a large number of different tools. In the \maze{} task, this is particularly prevalent, as fluids often become stuck stochastically in some funnels but not others of similar sizes (\autoref{fig:mpm_instability}).

Since the learned simulator was trained on much simpler scenarios where this effect is not observed, it only learns the smooth behavior of the fluid's movement, which makes the resulting trajectories look more realistic. This may enable better generalization to real world scenarios.

\subsubsection{Further designs found in \twodfluids{}}
\label{app:twodfluids-more-designs}
In the figures below, we demonstrate the range of found solutions for different solvers in tasks in the \twodfluids{} domain.

\begin{figure}[H]
\centering
\includegraphics[width=\textwidth]{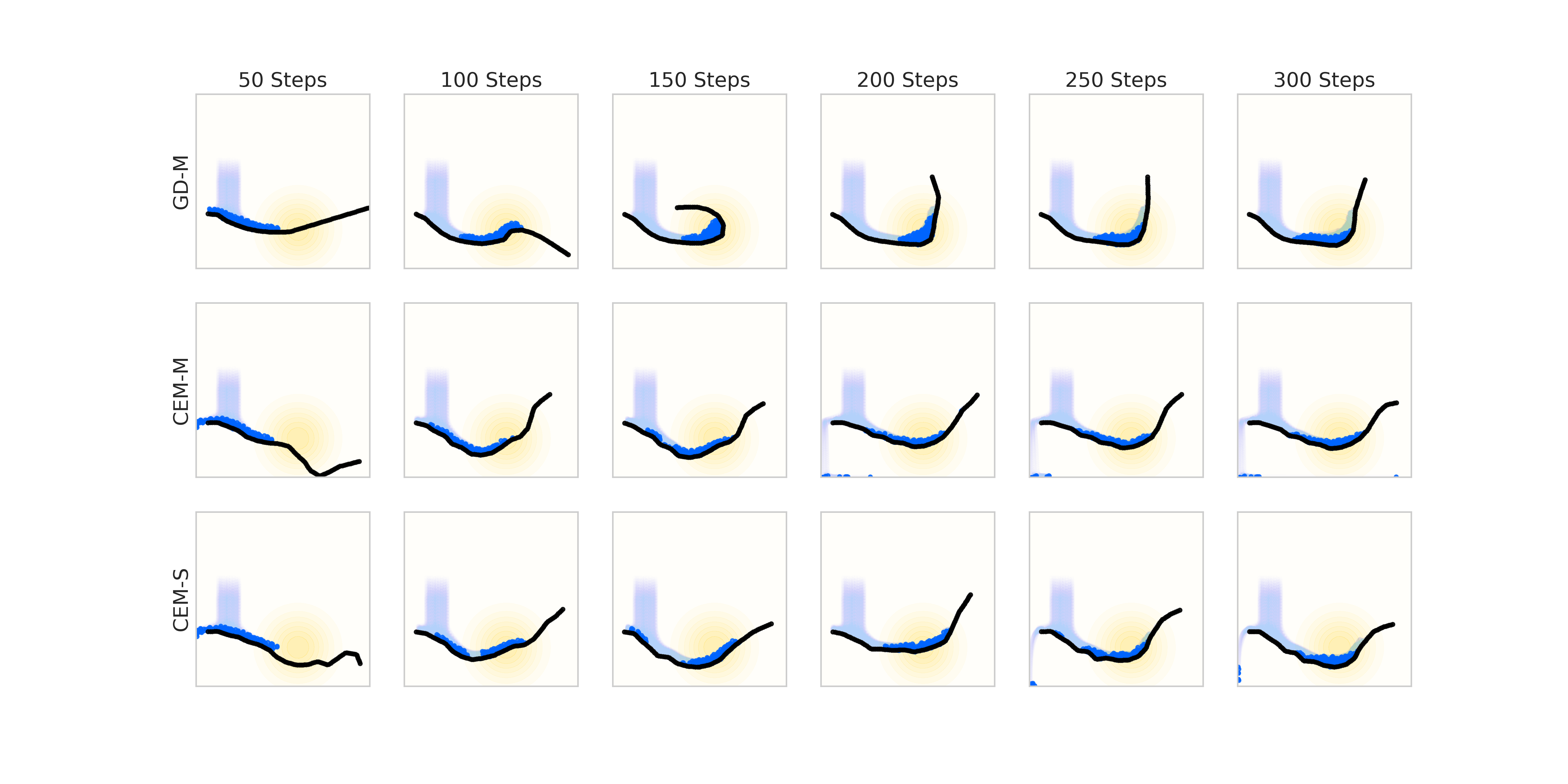}
\caption{Example solutions for each optimizer across the range of rollout lengths sampled for \contain{} in \autoref{fig:2dfluids_ablations}.}
\label{fig:hg_episode_sweep_demos}
\end{figure}

\begin{figure}[H]
\centering
\includegraphics[width=\textwidth]{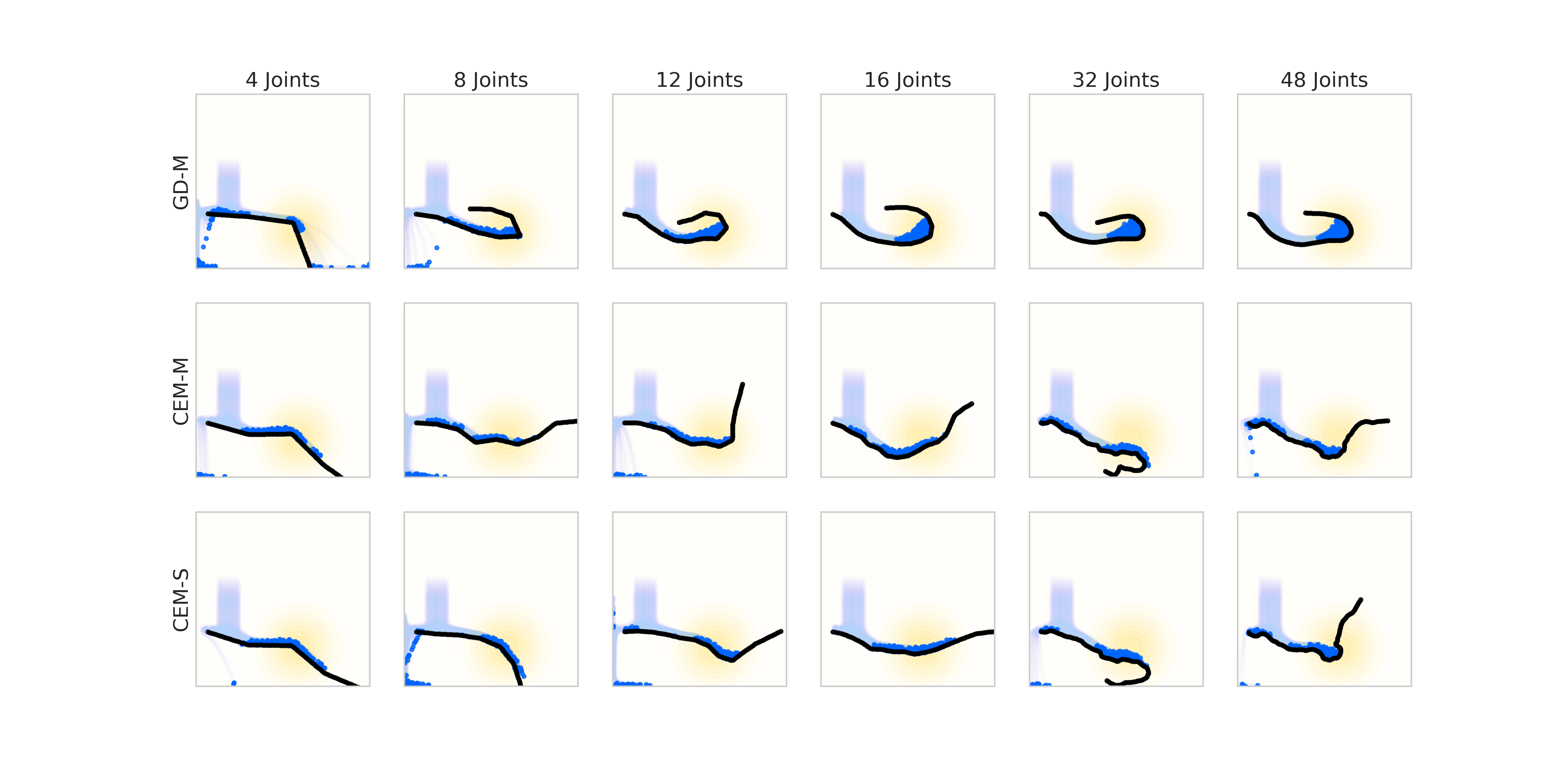}
\caption{Example solutions for each optimizer across the range of joint angle numbers sampled for \contain{} in \autoref{fig:2dfluids_ablations}.}
\label{fig:hg_n_joints_sweep_demos}
\end{figure}

\begin{figure}[H]
\centering
\includegraphics[width=.7\textwidth]{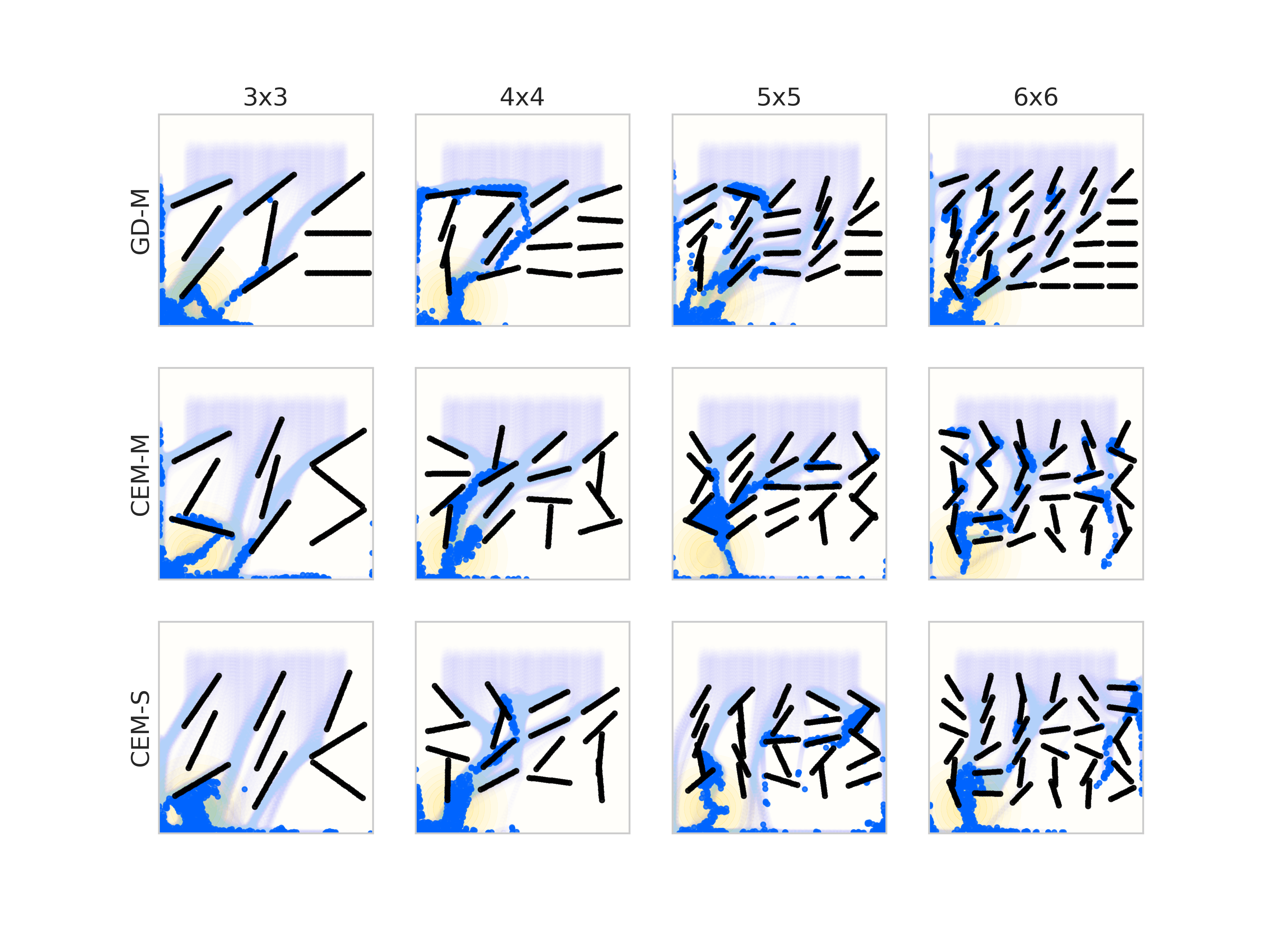}
\caption{Example solutions for each optimizer across the range of grid sizes sampled for \maze{} in \autoref{fig:2dfluids_ablations}.}
\label{fig:hg_maze_num_tools_demos}
\end{figure}

\begin{figure}[H]
\centering
\includegraphics[width=\textwidth]{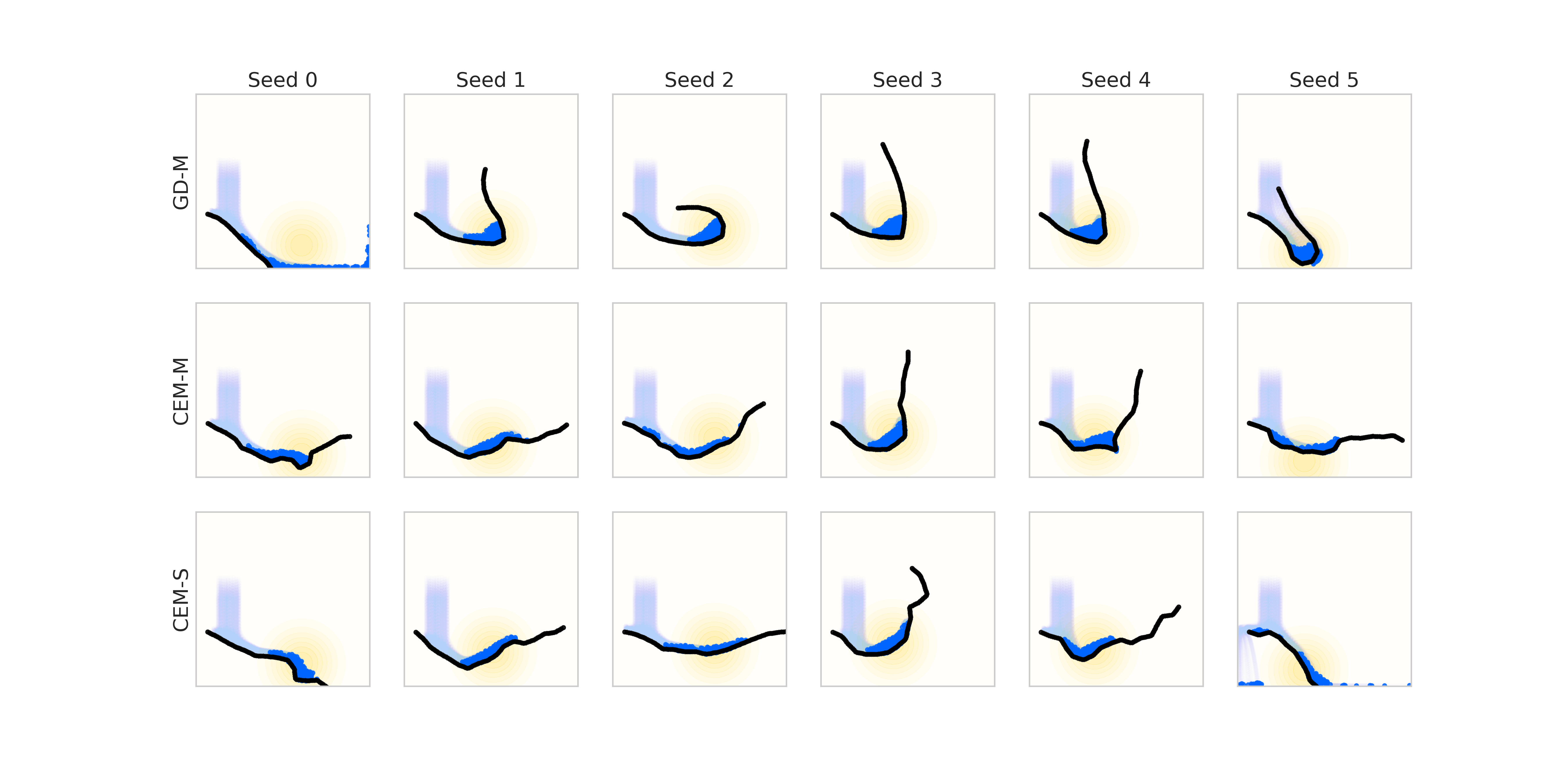}
\caption{Example solutions on \contain{} task for each optimizer across 6 random seeds.}
\label{fig:hg_contain_seeds_demos}
\end{figure}

\begin{figure}[H]
\centering
\includegraphics[width=\textwidth]{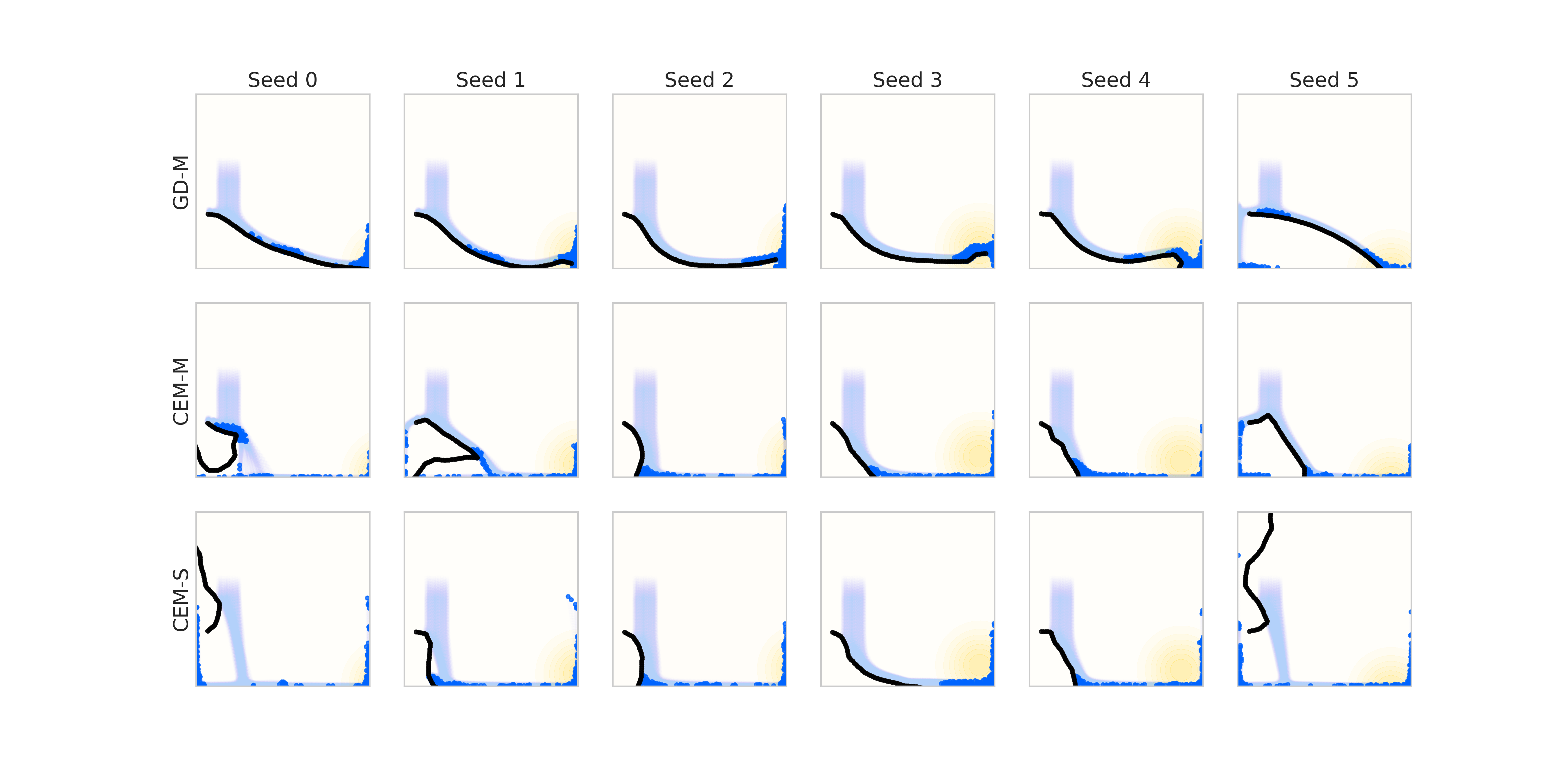}
\caption{Example solutions on \ramp{} task for each optimizer across 6 random seeds.}
\label{fig:hg_ramp_seeds_demos}
\end{figure}

\begin{figure}[H]
\centering
\includegraphics[width=\textwidth]{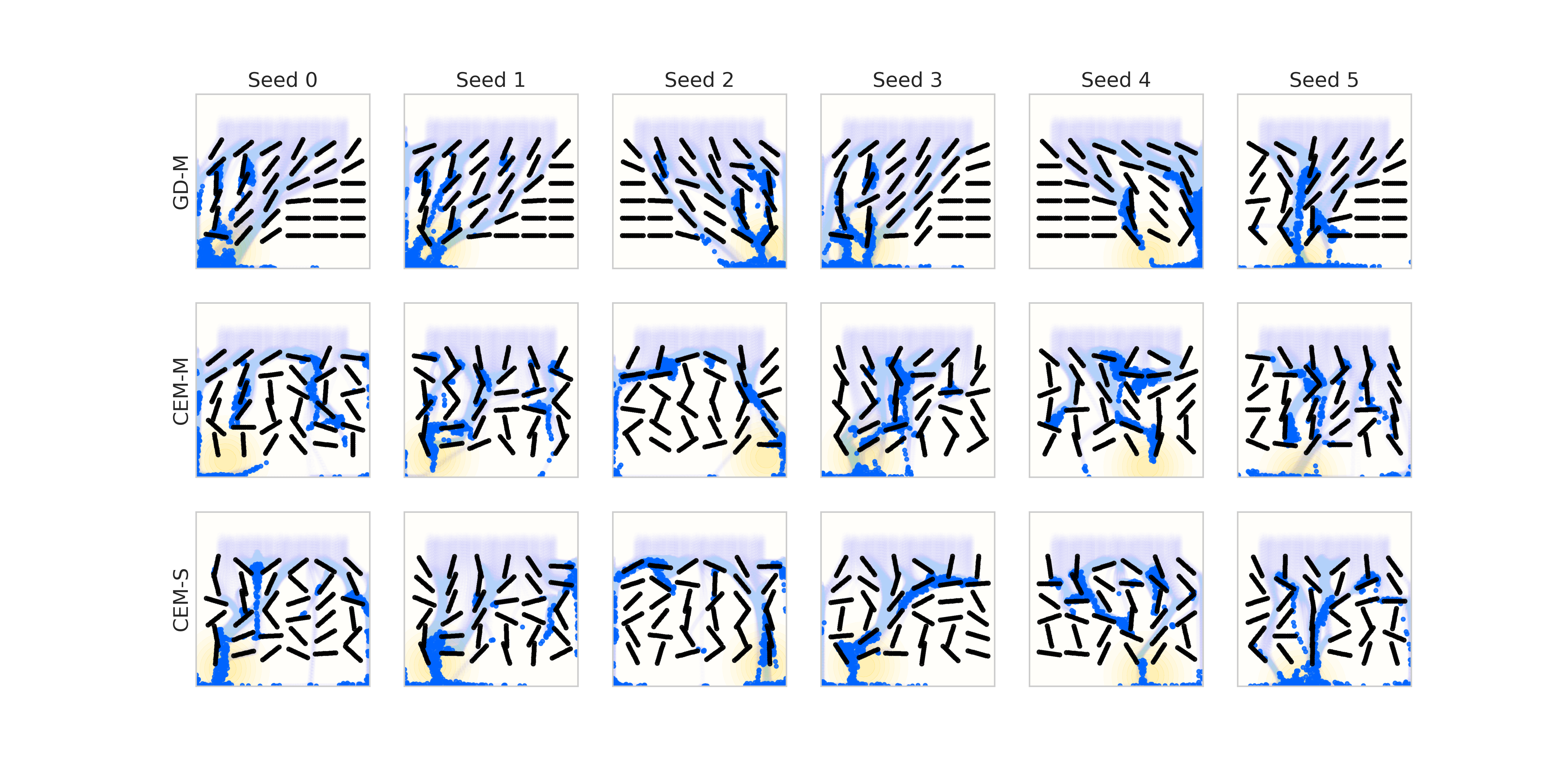}
\caption{Example solutions on \maze{} for each optimizer across 6 random seeds.}
\label{fig:hg_maze_seeds_demos}
\end{figure}

\end{document}